\documentclass[10pt,twocolumn]{article}

\usepackage[utf8]{inputenc}

\usepackage[T1]{fontenc}
\usepackage{lmodern}
\usepackage[letterpaper,top=0.72in,bottom=0.78in,left=0.68in,right=0.68in,columnsep=0.24in]{geometry}
\usepackage{amsmath,amsfonts}
\usepackage{algorithmic}
\usepackage{algorithm}
\usepackage{array}
\usepackage{textcomp}
\usepackage{stfloats}
\usepackage{url}
\usepackage{verbatim}
\usepackage{graphicx}
\usepackage{cite}
\usepackage[hidelinks]{hyperref}
\usepackage{rotating}
\usepackage{xcolor}
\usepackage[flushleft]{threeparttable}
\usepackage{booktabs}
\usepackage{multirow}
\usepackage{makecell}
\usepackage{microtype}
\usepackage{placeins}

\renewcommand{\topfraction}{0.95}
\renewcommand{\bottomfraction}{0.90}
\renewcommand{\textfraction}{0.05}
\renewcommand{\floatpagefraction}{0.85}
\begin{document}

\title{On the Adversarial Robustness of Remote Sensing Semantic\\Change Detection}

\author{Weikang Yu$^{1}$, Yonghao Xu$^{2,*}$, and Pedram Ghamisi$^{1}$\\[0.45em]
\small $^{1}$Helmholtz Institute Freiberg for Resource Technology,\\
\small Helmholtz-Zentrum Dresden-Rossendorf, 09599 Freiberg, Germany\\
\small $^{2}$Department of Electrical Engineering, Link{\"o}ping University,\\
\small 58183 Link{\"o}ping, Sweden\\
\small $^{*}$Corresponding author: Yonghao Xu (yonghaoxu@ieee.org)}

\date{\small arXiv preprint}
\maketitle

\begin{abstract}
Semantic change detection (SCD) is a bitemporal dense-prediction task that jointly identifies changed regions and their semantic states before and after change. Unlike single-image segmentation or binary change detection, SCD couples two temporal inputs with timestamp-wise semantic prediction, change localization, and final semantic-change decoding, creating adversarial dependencies that are not captured by conventional robustness protocols. We present a task-specific evaluation framework that separates output-side attack objectives from input-side temporal perturbation access, enabling systematic analysis of component vulnerability and cross-temporal propagation. Experiments on four datasets and six representative CNN-, Transformer-, and state-space-based models evaluate component-level and temporal objectives, single- and dual-timestamp perturbations, multiple attack methods, and cross-architecture transferability. The results show that final semantic-change predictions can be severely corrupted even when binary change localization remains comparatively stable, and that perturbations or attack objectives associated with one timestamp can propagate to the prediction of the other. These behaviors occur across different architecture families, while direct cross-model transfer remains considerably weaker than white-box attacks. The study demonstrates that adversarial robustness in SCD depends on the complete bitemporal prediction pathway rather than on an individual branch or backbone family, and provides a structured protocol for evaluating robustness in coupled bitemporal image analysis. Code is available at \href{https://github.com/EricYu97/AdvSCD}{https://github.com/EricYu97/AdvSCD}.
\end{abstract}

\par\medskip

\section{Introduction}
\label{sec1}

Semantic change detection (SCD) transforms bitemporal Earth observation imagery into detailed information about where land-cover changes occur and which semantic states are involved before and after each change \cite{cheng2024change}. Unlike binary change detection, which distinguishes only changed and unchanged regions, SCD characterizes explicit land-cover transitions between temporal observations. It therefore provides more actionable information for applications such as urban expansion monitoring \cite{yang2020second}, ecosystem disturbance analysis \cite{yuan2022transformer}, and agricultural monitoring \cite{wang2024cdsc-JL1}. As SCD models become increasingly integrated into automated image-analysis and decision-support workflows, their reliability under unexpected or deliberately manipulated inputs becomes an important concern.

Recent advances in deep learning have substantially improved SCD performance. Attention mechanisms \cite{ding2024joint-scannet}, Transformer architectures \cite{yuan2022transformer,long2025detecting-cpgnet}, and state-space models \cite{chen2024changemamba} have strengthened feature extraction, temporal interaction, and semantic-change reasoning. Nevertheless, most existing research evaluates these advances primarily through predictive accuracy on clean benchmark data. Strong clean performance does not establish whether a model will remain reliable under worst-case input disturbances. This limitation is particularly relevant to SCD because missed changes may conceal important events, false detections may generate unnecessary alarms, and incorrect semantic-transition labels may misrepresent the nature of environmental or anthropogenic change \cite{xu2023ai}. Robustness should therefore be assessed alongside clean accuracy when evaluating the reliability of SCD models.

\begin{figure*}[t]
\centering
\includegraphics[width=.7\linewidth]{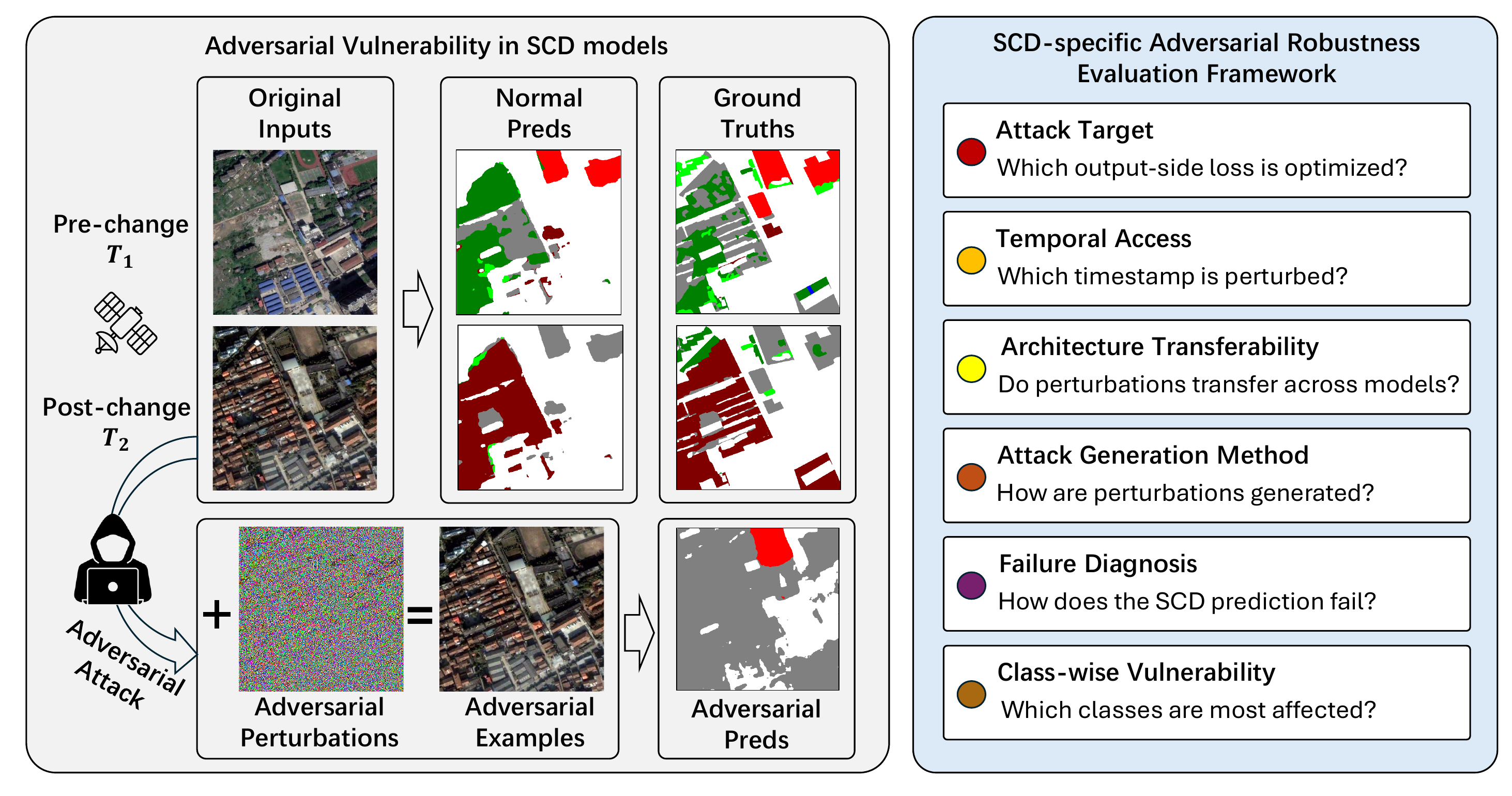}
\caption{Adversarial vulnerability and task-specific robustness evaluation in SCD. Norm-bounded perturbations affect timestamp-wise semantics, change localization, and semantic-change decoding. The framework evaluates attack objectives, temporal access, attack methods, transferability, and failure modes. Perturbations are magnified for visibility.}
\label{fig:intro_overview}
\end{figure*}

As illustrated in Fig.~\ref{fig:intro_overview}, adversarial attacks provide a controlled means of examining worst-case model behavior by deliberately optimizing small input perturbations that expose vulnerabilities and hidden dependencies not apparent under conventional evaluation \cite{javed2024robustness}. Such vulnerabilities have been investigated in image classification, object detection, and semantic segmentation, including their remote-sensing counterparts \cite{shi2022multifeature,zhang2022adversarial,bai2024stealthy}. More recently, adversarial analysis has been extended to binary change detection \cite{hao2025towards}. These studies demonstrate that adversarial robustness is determined not only by the backbone architecture but also by the structure of the prediction task and the way information is propagated through the model.

Existing adversarial evaluation protocols developed for semantic segmentation and binary change detection \cite{xie2017adversarial,gfanet,hao2025towards} do not adequately represent the coupled structure of SCD. Unlike single-image semantic segmentation or binary change detection, SCD integrates two temporal inputs, timestamp-wise semantic recognition, binary change localization, temporal interaction, and final semantic-change decoding \cite{cheng2024change}. This structure creates two complementary dimensions of vulnerability: output-side attack objectives targeting different prediction components and input-side perturbation access to either or both timestamps. Because adversarial effects may propagate across temporal pathways or be amplified when semantic and change predictions are combined, evaluating only one output or one temporal-access setting cannot fully characterize the adversarial behavior of SCD models.

To address this gap, we develop a task-specific framework that separates output-side attack objectives from input-side temporal perturbation access, enabling structured diagnosis of prediction coupling and cross-temporal propagation. We evaluate six representative CNN-, Transformer-, and Mamba-based models on four datasets using multiple perturbation methods, component-level and temporal objectives, and single- and dual-timestamp attacks. The evaluation further examines cross-architecture transferability, class-wise vulnerability, and three operationally distinct failure modes: change suppression, false-change hallucination, and semantic-transition corruption.

The main contributions of this work are summarized as follows:

\begin{itemize}
\item We formulate a task-specific adversarial robustness evaluation framework for SCD that explicitly accounts for bitemporal inputs, coupled prediction components, timestamp-specific outputs, and final semantic-change decoding. By separating output-side attack objectives from input-side temporal perturbation access, the framework enables structured diagnosis of both task-level and cross-temporal vulnerabilities.

\item We conduct a comprehensive evaluation across four datasets, six representative CNN-, Transformer-, and Mamba-based models, and multiple perturbation methods. The study covers component-level and temporal attack objectives, single- and dual-timestamp perturbations, cross-temporal propagation, and cross-architecture transferability.

\item We provide an SCD-specific interpretation of adversarial failure beyond aggregate performance degradation. The analysis identifies the final semantic-change prediction as the most effective average attack target and characterizes class- and dataset-dependent change suppression, false-change hallucination, and semantic-transition corruption, revealing how adversarial errors emerge in the coupled bitemporal prediction pipeline.

\end{itemize}

\section{Related Work}
\subsection{SCD in Remote Sensing}
\label{subsec:related_scd}

SCD extends binary change detection by jointly localizing changed regions and identifying their semantic states before and after change in co-registered multitemporal remote-sensing images \cite{cheng2024change}. By representing explicit land-cover transitions rather than only changed and unchanged areas, SCD supports more informative analysis of urban growth, environmental disturbance, agricultural dynamics, and land-use conversion. Early benchmarks such as HRSCD and SECOND primarily address semantic changes in high-resolution urban imagery \cite{daudt2019multitask-hrscd,yang2020second}. More recent datasets have broadened SCD to agricultural landscapes \cite{wang2024cdsc-JL1} and medium-resolution satellite observations \cite{lv2024semi-landsatscd}. Differences in spatial resolution, class distributions, and transition patterns across these benchmarks have driven the development of increasingly specialized SCD architectures.

Early deep SCD methods established the task as a coupled multi-output problem. HRSCD.str3 and HRSCD.str4 combine timestamp-wise land-cover mapping with binary change estimation through multitask learning \cite{daudt2019multitask-hrscd}. Subsequent work has focused on how semantic and change information should interact. Bi-SRNet uses semantic reasoning and consistency constraints to model intra-temporal and cross-temporal correlations \cite{ding2022bisrnet}, while ChangeMask exploits semantic-change causality and temporal symmetry within an encoder--Transformer--decoder architecture \cite{zheng2022changemask}. SCanNet further develops this direction through a CNN-based triple encoder--decoder and a Transformer variant that jointly models spatio-temporal semantic dependencies \cite{ding2024joint-scannet}. Collectively, these methods show that SCD performance depends not only on feature extraction, but also on the topology and strength of interactions among prediction branches.

Another line of research introduces structural priors and auxiliary supervision to improve semantic consistency and change localization. HGINet models relationships among bitemporal semantic and difference features through hierarchical graph interaction \cite{long2024semantic-hginet}, whereas CdSC combines cross-difference feature extraction with semantic co-alignment \cite{wang2024cdsc-JL1}. EGMS-Net uses coarse-to-fine enhancement and guidance modules to improve information exchange between semantic and change branches \cite{zuo2024multitask-egmsnet}. BGSNet adds boundary supervision to regularize changed-region geometry \cite{long2025bgsnet}, while CPGNet combines semantic-correlation modeling, a change-prior branch, and multi-view fusion \cite{long2025detecting-cpgnet}. These approaches reduce pseudo changes and semantic inconsistencies by constraining either intermediate representations or interactions among task-specific outputs.

Recent methods have expanded SCD beyond conventional CNN-based multitask designs. ChangeMamba introduces visual state-space modeling and dedicated spatio-temporal interaction modules for long-range dependency modeling \cite{chen2024changemamba}. FoBa combines a state-space backbone with foreground--background co-guidance and a finer-grained benchmark \cite{zhang2025foba}. GSTM-SCD extends state-space modeling from bitemporal to multitemporal semantic change analysis through graph-enhanced optimization and bidirectional 3D scanning \cite{liu2025gstm}, while Change3D reformulates change detection as tiny-video modeling with a unified video encoder \cite{zhu2025change3d}. These developments demonstrate growing diversity in how temporal evidence and semantic-change relationships are represented.

Despite this architectural diversity, existing SCD research has primarily optimized and compared performance under clean benchmark conditions. The same mechanisms that improve task interaction may also create pathways through which input perturbations propagate across timestamps, branches, and final outputs. Consequently, adversarial robustness cannot be inferred from backbone type or clean accuracy alone; it must be assessed at the level of the coupled SCD pipeline. This observation motivates evaluating how attack objectives and temporal perturbation access interact with different SCD designs.

\begin{figure*}[!htb]
    \centering
    \includegraphics[width=\linewidth]{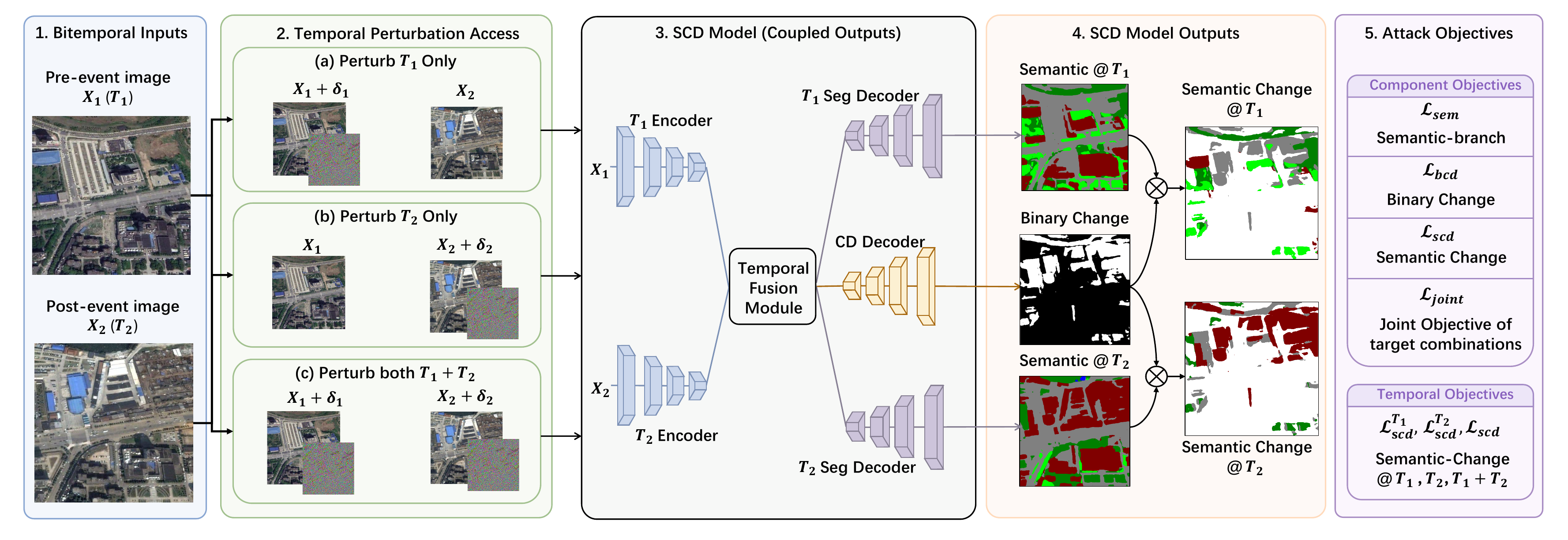}
    \caption{Task-specific adversarial robustness framework for SCD. Input access covers $T_1$, $T_2$, or both; output objectives target timestamp-wise semantics, binary change, or semantic change. Their combinations expose branch vulnerability, cross-temporal propagation, and final-output degradation.}
    \label{fig:overall_framework}
\end{figure*}

\subsection{Adversarial Attacks in Remote Sensing}
\label{subsec:related_adv_rs}


Research on adversarial vulnerability in remote sensing initially concentrated on single-image classification. Early studies demonstrated that scene classifiers can be misled by small perturbations and examined how vulnerability varies across models, datasets, and land-use categories \cite{czaja2018adversarial,xu2020assessing}. Subsequent work investigated attack selectivity and defensive strategies for remote-sensing scene recognition \cite{chen2020adversarialexampleremotesensing,chen2021empirical}. Adversarial analysis has also been extended to hyperspectral classification, where high-dimensional spectral information creates modality-specific attack surfaces \cite{sacnet}. For aerial object detection, attacks must disrupt both category recognition and spatial localization; existing work includes digital perturbations and adversarial patches \cite{huang2023adversarial,zhang2022adversarial}, as well as physically realizable attacks designed for overhead imagery \cite{du2022physical}. Together, these studies establish that remote-sensing models inherit adversarial vulnerabilities but express them through task- and sensor-dependent failure mechanisms.


Transferability is another important consideration when an attacker cannot access the target model. Mixup-Attack and Mixcut-Attack demonstrate that universal black-box perturbations can transfer across remote-sensing classifiers and semantic segmentation models \cite{xu2022uaers}. Targeted and source-targeted universal attacks further show that transferable manipulation can be achieved without querying the victim model during evaluation \cite{bai2022tuae}. More recently, pretrained models have been studied as surrogate models for attacking downstream remote-sensing systems \cite{bai2025transferadv}. Such findings motivate examining whether SCD architectures share transferable adversarial directions despite differences in their encoders and temporal-interaction mechanisms.

Compared with single-image interpretation, adversarial robustness in change detection remains less explored. Change detection depends on both the content of individual observations and their temporal correspondence, creating attack opportunities that do not arise in single-image tasks \cite{cheng2024change}. Recent work on remote-sensing building change detection has introduced generalizable physical attacks against bitemporal models, demonstrating that perturbations to one observation can compromise binary change localization \cite{hao2025towards}. However, this setting focuses on a single object category, binary changed--unchanged prediction, post-event perturbation, and patch-based physical threats. The robustness of change detection under norm-bounded attacks and alternative temporal-access settings therefore remains insufficiently characterized.

SCD further enlarges this attack surface by coupling pre-change and post-change semantic prediction, binary change localization, temporal interaction, and final semantic-change decoding. Contemporary SCD models commonly integrate these components through multitask \cite{daudt2019multitask-hrscd} or multibranch architectures \cite{zheng2022changemask}. Perturbations may consequently corrupt individual outputs, propagate between temporal pathways, or be amplified when semantic and change predictions are combined. Existing protocols for classification, segmentation, and binary change detection cannot distinguish these failure pathways. A task-specific evaluation must therefore consider both output-side attack objectives and input-side temporal perturbation access, while assessing the final semantic-change output and its underlying error modes. This unresolved need defines the gap addressed by our work.

\subsection{Trustworthy and Robust GeoAI}
\label{subsec:related_trustworthy_geoai}

Trustworthy GeoAI concerns whether geospatial AI systems operate reliably and responsibly in real-world applications \cite{ghamisi2025responsible,suhong2026operational}. Robustness is particularly important in Earth observation because model outputs may support environmental monitoring, disaster assessment, infrastructure management, and land-use analysis. Failures in these settings can produce missed events, false alarms, or incorrect interpretations of land-cover dynamics \cite{xu2023ai}. Adversarial evaluation provides a controlled way to expose such worst-case behavior.

Previous remote-sensing studies have examined adversarial attacks and defenses in classification and semantic segmentation \cite{gfanet,bai2024stealthy}, including diffusion-based input purification through UAD-RS \cite{yu2024universal}. Recent work on binary change detection has also shown that perturbations applied to one temporal observation can disrupt temporal comparison and change localization \cite{hao2025towards}. However, most existing protocols assume either a single input or one principal prediction output.

SCD presents a more complex robustness problem because its final prediction depends on two temporal inputs and several interacting outputs. A perturbation may affect semantic recognition, binary change localization, or final semantic-change decoding, while also propagating between timestamps. Existing evaluation protocols cannot clearly distinguish these pathways. A task-specific analysis of attack objectives, temporal perturbation access, and final prediction errors is therefore needed to characterize the adversarial reliability of SCD within trustworthy GeoAI.

\section{Task-Specific Adversarial Robustness Evaluation for SCD}
\label{sec:method}

This section presents a task-specific framework for evaluating adversarial robustness across the complete SCD pipeline. As shown in Fig.~\ref{fig:overall_framework}, the framework separates two complementary dimensions. The \textit{attack objective} specifies which output component is optimized: timestamp-wise semantic recognition, binary change localization, or final semantic-change decoding. The \textit{temporal access} specifies which input can be modified: the image at $T_1$, the image at $T_2$, or both. Their separation allows us to distinguish branch-level vulnerability, cross-temporal propagation, and degradation of the final SCD output within one evaluation protocol.

\subsection{SCD Task and Output Formulation}
\label{subsec:scd_formulation}

Let $X_1,X_2\in\mathbb{R}^{C_{\mathrm{in}}\times H\times W}$ be two co-registered images acquired at $T_1$ and $T_2$, and let $Y_1,Y_2\in\{1,\ldots,K\}^{H\times W}$ be their land-cover labels. All losses and metrics are computed only over the valid labeled pixels $\Omega_v$. The binary change label $Y_c$ indicates whether the two semantic labels differ. The final SCD label at each timestamp retains the land-cover class inside changed regions and assigns label $0$ to unchanged regions:
\begin{align} 
Y_c(p)&=\mathbb{I}[Y_1(p)\neq Y_2(p)], \nonumber\\ 
Y_{st}&=Y_t\odot Y_c,\qquad t\in\{1,2\}, 
\label{eq:scd_ground_truth_compact} 
\end{align} 
where $p\in\Omega_v$ and $\odot$ denotes pixel-wise multiplication. The binary mask $Y_c$ is broadcast over the semantic label map when needed. Thus, $Y_s=(Y_{s1},Y_{s2})$ represents the semantic states before and after each change, while unchanged pixels are assigned label $0$.

Because the evaluated models differ internally, we use a common output interface:
\begin{align}
&(S_1,S_2,B)=f_\theta(X_1,X_2), \nonumber\\
&\hat{Y}_t=\arg\max_k S_{t,k}, \nonumber \\
&\hat{Y}_c=\arg\max_m B_m, \nonumber\\
&S_{st}=S_t\odot B,\quad t\in\{1,2\}, \nonumber\\
&\hat{Y}_{st}=\Gamma(\hat{Y}_t,\hat{Y}_c).
\label{eq:scd_output_compact}
\end{align}
Here, $S_1$ and $S_2$ are the timestamp-wise semantic logits, while $B$ denotes the binary change-logit map. For models with a two-channel binary output, $B$ refers to the logit corresponding to the changed class. The binary change logits are broadcast across the semantic channels and multiplied element-wise with $S_t$ to obtain the final semantic-change logits $S_{st}$. These logits are used for attack optimization. For evaluation, the hard semantic-change prediction $\hat{Y}_{st}$ retains the semantic prediction $\hat{Y}_t$ within pixels predicted as changed by $\hat{Y}_c$ and assigns label $0$ elsewhere. One-channel binary outputs are handled using the equivalent binary formulation.

For transition-level analysis, the ordered pair $(Y_{s1}(p),Y_{s2}(p))$ is mapped to a unique label $Y_\Delta(p)$. A pair $(a,b)$ represents the transition $a\!\rightarrow\!b$, while $(0,0)$ denotes no change. This encoding is used only for transition-wise diagnosis; the two semantic-change maps remain the primary SCD output.

\subsection{Adversarial Threat Model}
\label{subsec:threat_model}

We consider untargeted test-time evasion attacks in the digital domain. Model parameters remain fixed. In the white-box setting, the attacker knows the model, preprocessing procedure, labels, and input gradients. Transfer-based black-box evaluation is described in Section~\ref{subsec:transfer_evaluation}.

Let $\mathcal{A}\subseteq\{1,2\}$ be the timestamps accessible to the attacker. Each accessible image receives an independent perturbation bounded by $\epsilon$ in the original $[0,1]$ image space. For a selected attack objective $\mathcal{L}_{\mathrm{atk}}$, the adversarial example is obtained from
\begin{align}
\{\delta_t^*\}_{t\in\mathcal{A}}
&=\arg\max_{\|\delta_t\|_\infty\leq\epsilon,\;t\in\mathcal{A}}
\mathcal{L}_{\mathrm{atk}}
\!\left(f_\theta(X_1^{\mathrm{adv}},X_2^{\mathrm{adv}}),Y\right),
\nonumber\\
X_t^{\mathrm{adv}}
&=\begin{cases}
\Pi_{\mathcal{X}}(X_t+\delta_t),&t\in\mathcal{A},\\
X_t,&t\notin\mathcal{A},
\end{cases}
\label{eq:threat_model_compact}
\end{align}
where $\Pi_{\mathcal{X}}$ clips an image to its valid range and $Y$ denotes the labels required by the selected objective. Invalid pixels are excluded from optimization. Under joint access, $\delta_1$ and $\delta_2$ each satisfy the same budget rather than sharing a combined budget.

For models using channel-wise normalization, the valid bounds and perturbation budget are converted with the dataset mean and standard deviation before optimization. Projection onto both the perturbation set and the valid input range is applied after every update.

\subsection{Attack Objectives in SCD}
\label{subsec:attack_targets}

The coupled outputs of an SCD model support several complementary attack objectives. We use cross-entropy for the timestamp-wise semantic branches, the binary-change branch, and the two final semantic-change maps:
\begin{align}
\mathcal{L}_{\mathrm{sem}}
&=\mathcal{L}_{\mathrm{ce}}(S_1,Y_1)
+\mathcal{L}_{\mathrm{ce}}(S_2,Y_2), \nonumber\\
\mathcal{L}_{\mathrm{bcd}}
&=\mathcal{L}_{\mathrm{ce}}(B,Y_c), \nonumber\\
\mathcal{L}_{\mathrm{scd}}^{T_t}
&=\mathcal{L}_{\mathrm{ce}}(S_{st},Y_{st}), \nonumber \\
\mathcal{L}_{\mathrm{scd}}
&=\mathcal{L}_{\mathrm{scd}}^{T_1}
+\mathcal{L}_{\mathrm{scd}}^{T_2}.
\label{eq:attack_objectives_compact}
\end{align}
Binary cross-entropy with logits is used when a model has a one-channel binary output. The semantic objective tests whether corrupting land-cover recognition alone propagates to the final result; the binary objective isolates change-localization vulnerability; and the semantic-change objective directly optimizes the output used for the primary SCD evaluation.

The component objectives can also be combined as
\begin{equation}
\mathcal{L}_{\mathrm{joint}}
=\lambda_{\mathrm{sem}}\mathcal{L}_{\mathrm{sem}}
+\lambda_{\mathrm{bcd}}\mathcal{L}_{\mathrm{bcd}}
+\lambda_{\mathrm{scd}}\mathcal{L}_{\mathrm{scd}}.
\label{eq:joint_objective_compact}
\end{equation}
The semantic, binary-change, semantic-change, semantic+binary-change, and complete joint targets use weights $(1,0,0)$, $(0,1,0)$, $(0,0,1)$, $(1,1,0)$, and $(1,1,1)$, respectively. All active terms use the same valid-pixel reduction.

In addition to these component-level objectives, we evaluate temporal semantic-change targets. Optimizing only $\mathcal{L}_{\mathrm{scd}}^{T_1}$ or $\mathcal{L}_{\mathrm{scd}}^{T_2}$ attacks one side of the final output, whereas optimizing their sum attacks both. This output-side choice is independent of temporal perturbation access: an attack may optimize one temporal output while modifying either input or both.

\subsection{Temporal Perturbation Access}
\label{subsec:temporal_access}

We consider three input-side access settings. Under $T_1$-only access, $\mathcal{A}=\{1\}$ and the later image remains clean. Under $T_2$-only access, $\mathcal{A}=\{2\}$ and the reference image remains clean. Under joint bitemporal access, $\mathcal{A}=\{1,2\}$ and both images receive independently constrained perturbations.

The single-timestamp settings test whether manipulating one observation is sufficient to affect predictions associated with both times. Their comparison reveals temporal asymmetry, while joint access provides the strongest input-side setting considered in this work. In every case, the loss is computed from the complete model output, but gradients update only the accessible inputs.

\subsection{Attack Generation Methods}
\label{subsec:attack_generation}

We instantiate the objectives and access settings with two non-optimized baselines and four gradient-based attacks. Gaussian noise is sampled independently, clipped to $[-\epsilon,\epsilon]$, and added to each accessible input. Jitter applies the predefined small spatial translation used in our experiments. These baselines do not optimize against the model and therefore distinguish ordinary input sensitivity from adversarial vulnerability.

FGSM, I-FGSM, and PGD use the gradient of the selected loss with respect to each accessible perturbation. Their common update is
\begin{equation}
\delta_t^{r+1}
=\Pi_{\epsilon,\mathcal{X}}
\left[
\delta_t^r+\alpha\,
\operatorname{sign}
\left(\nabla_{\delta_t}
\mathcal{L}_{\mathrm{atk}}\right)
\right],
\qquad t\in\mathcal{A},
\label{eq:gradient_update_compact}
\end{equation}
where $\Pi_{\epsilon,\mathcal{X}}$ enforces both the $\ell_\infty$ budget and the valid image range. FGSM takes one step with $\alpha=\epsilon$. I-FGSM starts from zero and applies repeated steps, whereas PGD starts from a random point in $[-\epsilon,\epsilon]$ before applying the same iterative update. PGD is our default white-box attack because it provides a strong iterative test under the stated threat model.

We also include a C\&W-style margin attack on the two semantic-change outputs. For each valid pixel and temporal output, it penalizes the margin between the ground-truth logit and the strongest incorrect logit:
\begin{align}
m_t(p)
&=
S_{st}^{\,Y_{st}(p)}(p)
-\max_{k\neq Y_{st}(p)}S_{st}^{\,k}(p)
+\kappa, \nonumber\\
\mathcal{L}_{\mathrm{cw}}
&=\frac{1}{|\Omega_v|}
\sum_{p\in\Omega_v}\sum_{t=1}^{2}
\max[m_t(p),0].
\label{eq:cw_compact}
\end{align}
Because our attack procedure performs gradient ascent, we set $\mathcal{L}_{\mathrm{atk}}=-\mathcal{L}_{\mathrm{cw}}$ as the objective. This confidence-based loss complements the cross-entropy attacks without changing the temporal-access constraints.

\subsection{Robustness Metrics and Failure Diagnosis}
\label{subsec:metrics_diagnosis}

We report the standard SCD metrics used by the evaluated benchmarks: overall accuracy, mIoU, SeK, binary change F-score $F_{\mathrm{bcd}}$, foreground semantic F-score $F_{\mathrm{seg}}$, and semantic change F-score $F_{\mathrm{scd}}$. The binary score collapses all semantic change classes into one changed class. The foreground semantic score averages the timestamp-wise land-cover F-scores. The semantic-change score counts a changed pixel as correct only when both its change status and semantic category are correct. We therefore use $F_{\mathrm{scd}}$ as the primary performance measure, with $F_{\mathrm{bcd}}$ and $F_{\mathrm{seg}}$ diagnosing whether degradation originates mainly from change localization or semantic recognition. mIoU and SeK are retained for compatibility with prior SCD work.

Attack success is measured by relative F-score degradation:
\begin{equation}
\mathrm{ASR}_m
=\frac{F_m^{\mathrm{clean}}-F_m^{\mathrm{adv}}}
{F_m^{\mathrm{clean}}+\varepsilon},
\qquad
m\in\{\mathrm{bcd},\mathrm{seg},\mathrm{scd}\}.
\label{eq:asr_general}
\end{equation}
Here, $\varepsilon$ is a small numerical constant used only to avoid division by zero. $\mathrm{ASR}_{\mathrm{scd}}$ is the primary robustness indicator. We do not derive attack success from SeK because this Kappa-based measure can be negative.

Aggregate scores do not reveal how an attack changes the output. We therefore define three task-specific diagnostic rates. Let $\hat{Y}_c$ and $\hat{Y}_c^{\mathrm{adv}}$ be the clean and adversarial binary predictions, and let $\hat{Y}_\Delta^{\mathrm{adv}}$ be the adversarial transition prediction. Using $\#\{\cdot\}$ to count valid pixels satisfying a condition, we compute
\begin{align}
R_{\mathrm{sup}}
&=\frac{\#\{Y_c=1,\hat{Y}_c=1,\hat{Y}_c^{\mathrm{adv}}=0\}}
{\#\{Y_c=1,\hat{Y}_c=1\}+\varepsilon},
\nonumber\\
R_{\mathrm{hal}}
&=\frac{\#\{Y_c=0,\hat{Y}_c^{\mathrm{adv}}=1\}}
{\#\{Y_c=0\}+\varepsilon},
\nonumber\\
R_{\mathrm{trans}}
&=\frac{\#\{Y_c=1,\hat{Y}_c^{\mathrm{adv}}=1,
\hat{Y}_\Delta^{\mathrm{adv}}\neq Y_\Delta\}}
{\#\{Y_c=1,\hat{Y}_c^{\mathrm{adv}}=1\}+\varepsilon}.
\label{eq:failure_rates_compact}
\end{align}
$R_{\mathrm{sup}}$ measures real changes detected by the clean model but hidden by the attack; $R_{\mathrm{hal}}$ measures false changes introduced in unchanged areas; and $R_{\mathrm{trans}}$ measures incorrect semantic transitions among ground-truth changes that remain detected. Together, they separate concealment, false alarms, and semantic misinterpretation.

\subsection{Cross-Architecture Transferability}
\label{subsec:transfer_evaluation}

Finally, we evaluate whether perturbations generated on one architecture transfer to another. A source model $f_{\theta_s}$ is used to optimize the perturbations under the same objective, budget, and temporal-access constraints defined above. The resulting adversarial image pair is then passed directly to a target model $f_{\theta_t}$ without using its gradients. Identical source and target models yield the white-box diagonal of the transfer matrix; different models yield the off-diagonal transfer-based black-box results. This analysis indicates whether the attacks exploit model-specific behavior or weaknesses shared across SCD architectures and temporal-fusion designs.

\section{Experiments and Results}
\label{sec:experiments}

This section describes the experimental setup and reports the corresponding robustness results. After establishing clean-performance baselines, we evaluate the default white-box setting and then vary one protocol dimension at a time: attack objective, temporal target, perturbation access, attack generator, and cross-architecture transferability. We finally report error-mode and class-wise diagnostics. The main paper presents the aggregate comparisons and the principal error-mode visualization; dataset--model breakdowns, class-wise results, and additional qualitative examples are provided in the supplementary material. Cross-cutting trends and their broader implications are discussed in Section~\ref{sec:discussion}.

\subsection{Datasets}
\label{subsec:datasets}

Experiments are conducted on four SCD datasets: HiUCD-Mini, JL1, LandsatSCD, and SECOND. As summarized in Table~\ref{tab:datasets}, these datasets cover different spatial resolutions, scene types, image sizes, and semantic label systems, providing a diverse benchmark for evaluating adversarial vulnerability across SCD scenarios. Each sample consists of a pair of co-registered remote-sensing images acquired at two timestamps, together with semantic annotations for the pre-event and post-event images. Following the SCD formulation in Section~\ref{subsec:scd_formulation}, binary change labels and final semantic change labels are derived from the timestamp-wise semantic annotations, while invalid or unlabeled pixels are excluded from both attack optimization and evaluation. 





\begin{table*}[t]
    \centering
    \caption{Evaluated SCD datasets. Class counts include the no-change/no-label class.}
    \label{tab:datasets}
    \small
    \setlength{\tabcolsep}{12pt}
    \begin{tabular*}{\linewidth}{@{\extracolsep{\fill}}lcccccc@{}}
        \toprule
        Dataset & Image size & Resolution & Channels & Classes & Scene type & \makecell{Dataset split (Patches)\\ Train/Val./Test} \\
        \midrule
        HiUCD-Mini \cite{tian2020hiucd} & $256 \times 256$ & 0.1 m & 3 (RGB) & 10 & Urban & 4800/944/6176 \\
        JL1 \cite{wang2024cdsc-JL1} & $256 \times 256$ & 0.75 m & 3 (RGB) & 6 & Cropland & 4050/1950/1950 \\
        LandsatSCD \cite{yuan2022transformer} & $416 \times 416$ & 30 m & 3 (RGB) & 5 & Dryland & 1431/477/477 \\
        SECOND \cite{yang2020second} & $512 \times 512$ & 0.5--3 m & 3 (RGB) & 7 & Urban & 2968/1694/1694 \\
        \bottomrule
    \end{tabular*}
\end{table*}


\subsection{Evaluated Models}
\label{subsec:models}

To examine how adversarial vulnerability varies across architectural designs, we evaluate six SCD models grouped by their principal encoder family: CNN, Transformer, and visual state-space (Mamba) models. Two models are selected from each group, covering different temporal interaction mechanisms, model capacities, and computational costs.

\begin{table*}[t]
    \centering
    \caption{SCD models evaluated in the main experiments. FLOPs are for one $512 \times 512$ bitemporal image pair.}
    \label{tab:models}
    \small
    \setlength{\tabcolsep}{12pt}
    \begin{tabular*}{\linewidth}{@{\extracolsep{\fill}}lccccc@{}}
        \toprule
        Model & Family & Encoder & Fusion & Params. (M) & FLOPs (G) \\
        \midrule
        HRSCD4 \cite{daudt2019multitask-hrscd} & CNN & UNet-like & Concat. & 13.71 & 36.73 \\
        EGMSNet \cite{zuo2024multitask-egmsnet} & CNN & ResNet-34 & Diff. & 30.73 & 215.61 \\
        BGSNet \cite{long2025bgsnet} & Transformer & PVTv2-B2 & Cross-Attn.+Diff. & 25.18  & 117.05 \\
        CPGNet \cite{long2025detecting-cpgnet} & Transformer & PVTv2-B2 & Diff.+Concat. & 55.38 & 87.24 \\
        ChangeMamba \cite{chen2024changemamba} & Mamba & VSSM-Base & SSM+Concat. & 97.04 & 226.30 \\
        FoBa \cite{zhang2025foba} & Mamba & VSSM-Small & Mul.+Concat. & 53.95 & 107.33 \\
        \bottomrule
    \end{tabular*}
\end{table*}

 As summarized in Table~\ref{tab:models}, the selected models cover different inductive biases and temporal interaction mechanisms. HRSCD4 and EGMSNet represent convolutional architectures, where temporal information is mainly combined through feature concatenation or differencing. BGSNet and CPGNet use Transformer-based encoders to capture broader spatial context and model temporal interaction. ChangeMamba and FoBa represent recent state-space-based architectures for long-range dependency modeling. This model set allows us to analyze whether adversarial vulnerability is tied to specific architectural choices or appears consistently across different SCD designs.

All models are trained and evaluated under their SCD prediction settings. During adversarial evaluation, model parameters are fixed and all models are set to evaluation mode. Attacks are applied only at test time, and the same perturbation budgets, attack targets, temporal perturbation access settings, and attack generation methods are used across models.

\subsection{Implementation Details}
\label{subsec:implementation_details}
PGD is the default white-box optimizer, with $\epsilon=2/255$, $\alpha=0.5/255$, and 10 iterations. All norm-bounded comparison attacks use $\epsilon=2/255$. FGSM performs a single update, whereas I-FGSM uses $\alpha=0.5/255$ for $10$ iterations. The C\&W-style attack maximizes Eq.~\eqref{eq:cw_compact} for $20$ Adam iterations using a learning rate of $10^{-3}$, $\beta=0.01$, and a confidence margin of $\kappa=0$. Gaussian noise uses $\sigma=0.5/255$ and is clipped to the same perturbation bound. Spatial jitter applies independently sampled horizontal and vertical displacements from $\{-1,0,1\}$ pixels to each temporal image. These settings are fixed across datasets and models unless otherwise specified and are used as a common operating point rather than claimed to be globally optimal.

The default protocol uses the semantic-change objective with joint access to $T_1$ and $T_2$. Subsequent experiments vary one protocol axis at a time. The comparison objectives use the unit-weight combinations defined in Section~\ref{subsec:attack_targets}.

All attacks are generated on the test sets with model parameters fixed. Invalid or unlabeled pixels are excluded from optimization and evaluation, and dataset-specific normalization is accounted for when enforcing the input bounds.

Experiments are implemented in PyTorch and conducted on an HPC system equipped with NVIDIA A100 GPUs. Training and adversarial evaluation use identical dataset-specific preprocessing, and all reported adversarial results are computed on the corresponding test splits unless otherwise specified.
\subsection{Clean SCD Performance}
\label{subsec:clean_performance}

Table~\ref{tab:clean_performance} reports the clean test-set performance averaged over the four datasets; Supplementary Table~S1 provides the dataset--model breakdown. These model-level averages establish the non-adversarial baselines used to compute relative performance degradation under attack. Clean and adversarial metrics are calculated from the same trained checkpoints, test splits, preprocessing pipeline, and output-decoding procedure.

\begin{table}[t]
    \centering
    \caption{Clean performance of the evaluated models averaged on four SCD datasets (\%).}
    \label{tab:clean_performance}
    \resizebox{\linewidth}{!}{
    \begin{tabular}{l|cccccc}
        \toprule
        Model & OA & $P_{\mathrm{scd}}$ & $R_{\mathrm{scd}}$ 
        & $F_{\mathrm{scd}}$ & mIoU & SeK \\
        \midrule
        HRSCD4     & 89.17 & 60.79 & 62.36 & 61.51 & 72.85 & 22.40 \\
        EGMSNet    & 92.97 & 74.48 & 74.23 & 74.28 & 80.26 & 40.46 \\
        BGSNet     & 91.09 & 78.19 & 65.08 & 70.57 & 76.58 & 33.82 \\
        CPGNet     & 93.24 & 74.89 & 75.75 & 75.15 & 80.45 & 40.84 \\
        ChangeMamba & 93.81 & 74.51 & 77.72 & 76.00 & 81.40 & 42.66 \\
        FoBa       & 93.42 & 70.83 & 76.58 & 73.50 & 79.59 & 39.12 \\
        \bottomrule
    \end{tabular}
    }
\end{table}

\begin{table*}[t]
\centering
\caption{Robustness under the default white-box setting: joint bitemporal PGD with the semantic-change target. Attack success rates are relative F-score reductions. The values are the average of results from four datasets.}
\label{tab:standard_pgd_robustness}
\footnotesize
\setlength{\tabcolsep}{3pt}
\begin{tabular*}{\linewidth}{l@{\hspace{6pt}}|@{\extracolsep{\fill}}ccc|ccc|ccc}
\toprule
Model
& $F_{\mathrm{bcd}}^{clean}$
& $F_{\mathrm{bcd}}^{adv}$
& $\mathrm{ASR}_{\mathrm{bcd}}$
& $F_{\mathrm{seg}}^{clean}$
& $F_{\mathrm{seg}}^{adv}$
& $\mathrm{ASR}_{\mathrm{seg}}$
& $F_{\mathrm{scd}}^{clean}$
& $F_{\mathrm{scd}}^{adv}$
& $\mathrm{ASR}_{\mathrm{scd}}$ \\
\midrule
HRSCD4      & 71.39 & 52.98 & 25.46 & 29.00 & 12.28 & 57.06 & 61.51 & 10.77 & 85.36 \\
EGMSNet     & 80.12 & 60.85 & 23.11 & 46.48 & 14.55 & 70.57 & 74.28 &  8.36 & 89.62 \\
BGSNet      & 76.60 & 52.25 & 31.76 & 44.66 & 13.17 & 73.52 & 70.57 &  8.01 & 89.20 \\
CPGNet      & 80.12 & 57.48 & 27.63 & 42.23 & 15.41 & 68.25 & 75.15 &  8.15 & 89.59 \\
ChangeMamba & 80.95 & 57.77 & 27.67 & 45.25 & 16.57 & 66.97 & 76.00 &  8.29 & 89.69 \\
FoBa        & 78.40 & 58.31 & 23.45 & 46.10 & 13.25 & 71.39 & 73.50 & 11.27 & 85.20 \\
\bottomrule
\end{tabular*}
\end{table*}

In Table~\ref{tab:clean_performance}, averaged $F_{\mathrm{scd}}$ ranges from 61.51 to 76.00. ChangeMamba leads in $F_{\mathrm{scd}}$, mIoU (81.40), and SeK (42.66), while BGSNet has the highest precision (78.19). These clean values are the denominators for the model-level attack-success rates below.
\subsection{Main Robustness under the Default White-Box Setting}
\label{subsec:standard_pgd_robustness}

We first evaluate robustness under the default white-box attack. PGD jointly perturbs both temporal inputs using the semantic-change objective $\mathcal{L}_{\mathrm{scd}}$. This task-aligned setting directly optimizes degradation of the final SCD output while allowing the perturbation to modify the complete bitemporal input pair.

Table~\ref{tab:standard_pgd_robustness} reports the model-level averages; Supplementary Table~S2 provides the dataset--model breakdown. Under attack, averaged $F_{\mathrm{scd}}$ falls to 8.01--11.27, and $\mathrm{ASR}_{\mathrm{scd}}$ reaches 85.20--89.69\% for all six models. The corresponding ranges are lower for $\mathrm{ASR}_{\mathrm{bcd}}$ (23.11--31.76\%) and $\mathrm{ASR}_{\mathrm{seg}}$ (57.06--73.52\%). Thus, the final semantic-change output is consistently degraded more strongly than binary change localization, establishing the default baseline for the controlled comparisons below.

\subsection{Attack Target Analysis}
\label{subsec:attack_target_results}

We next analyze how different output-side attack targets affect SCD robustness.

\subsubsection{Component-Level Attack Targets}
\label{subsubsec:component_target_results}
To examine how objective selection shapes adversarial failure, we fix perturbation access to the joint bitemporal setting and use PGD as the attack generator. We compare five untargeted loss objectives: semantic-branch, binary-change, semantic+binary-change, semantic-change, and joint objectives. These objectives optimize timestamp-wise semantic recognition, binary change localization, both component predictions, the final SCD output, and all available outputs, respectively.

Table~\ref{tab:objective_comparison_diagnostic} summarizes the component-level objectives; Supplementary Table~S3 gives the dataset-specific results. The binary-change objective gives the largest $\mathrm{ASR}_{\mathrm{bcd}}$ for every model average (70.20--85.57\%), whereas the semantic-change objective gives the largest $\mathrm{ASR}_{\mathrm{scd}}$ for five of six models (85.20--89.69\%) while retaining a much lower $\mathrm{ASR}_{\mathrm{bcd}}$ (23.11--31.76\%). The two objectives therefore emphasize different components of the prediction pipeline.

\begin{table*}[t]
\centering
\caption{Attack success rates for component-level targets under joint bitemporal PGD. The values are the average of results from four datasets. }
\label{tab:objective_comparison_diagnostic}
\scriptsize
\setlength{\tabcolsep}{2.2pt}
\renewcommand{\arraystretch}{0.95}
\resizebox{\linewidth}{!}{
\begin{tabular}{l|ccc|ccc|ccc|ccc|ccc}
\toprule
\multirow{2}{*}{Model}
& \multicolumn{3}{c|}{Sem.}
& \multicolumn{3}{c|}{BCD}
& \multicolumn{3}{c|}{Sem.+BCD}
& \multicolumn{3}{c|}{SCD}
& \multicolumn{3}{c}{Joint} \\
\cmidrule(lr){2-4}
\cmidrule(lr){5-7}
\cmidrule(lr){8-10}
\cmidrule(lr){11-13}
\cmidrule(lr){14-16}
& $\mathrm{ASR}_{\mathrm{bcd}}$
& $\mathrm{ASR}_{\mathrm{seg}}$
& $\mathrm{ASR}_{\mathrm{scd}}$
& $\mathrm{ASR}_{\mathrm{bcd}}$
& $\mathrm{ASR}_{\mathrm{seg}}$
& $\mathrm{ASR}_{\mathrm{scd}}$
& $\mathrm{ASR}_{\mathrm{bcd}}$
& $\mathrm{ASR}_{\mathrm{seg}}$
& $\mathrm{ASR}_{\mathrm{scd}}$
& $\mathrm{ASR}_{\mathrm{bcd}}$
& $\mathrm{ASR}_{\mathrm{seg}}$
& $\mathrm{ASR}_{\mathrm{scd}}$
& $\mathrm{ASR}_{\mathrm{bcd}}$
& $\mathrm{ASR}_{\mathrm{seg}}$
& $\mathrm{ASR}_{\mathrm{scd}}$ \\
\midrule
HRSCD4      & 18.40 & 58.62 & 59.06 & 70.20 & 24.05 & 73.33 & 61.12 & 55.85 & 71.30 & 25.46 & 57.06 & 85.36 & 53.67 & 64.31 & 81.27 \\
EGMSNet     & 44.07 & 57.79 & 66.08 & 85.57 & 60.15 & 88.60 & 45.81 & 58.15 & 67.01 & 23.11 & 70.57 & 89.62 & 44.24 & 67.12 & 77.78 \\
BGSNet      & 49.94 & 67.75 & 68.80 & 72.28 & 51.50 & 76.24 & 63.73 & 63.53 & 72.46 & 31.76 & 73.52 & 89.20 & 53.07 & 72.58 & 77.95 \\
CPGNet      & 35.68 & 57.39 & 60.70 & 75.53 & 48.79 & 78.92 & 60.50 & 55.31 & 69.50 & 27.63 & 68.25 & 89.59 & 55.03 & 66.98 & 80.91 \\
ChangeMamba & 46.51 & 73.82 & 71.96 & 79.69 & 55.37 & 83.45 & 72.09 & 73.29 & 78.89 & 27.67 & 66.97 & 89.69 & 55.23 & 77.38 & 82.71 \\
FoBa        & 46.74 & 71.34 & 74.81 & 85.18 & 67.85 & 87.47 & 67.47 & 72.98 & 80.74 & 23.45 & 71.39 & 85.20 & 52.94 & 80.24 & 84.74 \\
\bottomrule
\end{tabular}
}
\vspace{1mm}
\begin{tablenotes}
\footnotesize
\item Sem., BCD, SCD, and Joint denote attacks optimized using $(\lambda_{sem},\lambda_{bcd},\lambda_{scd})=(1,0,0)$, $(0,1,0)$, $(0,0,1)$, and $(1,1,1)$, respectively. Sem.+BCD corresponds to $(1,1,0)$.
\end{tablenotes}
\end{table*}

\begin{table*}[!htb]
    \centering
    \caption{Attack success rates for temporal semantic-change targets under joint bitemporal PGD. The loss targets the $T_1$ output, the $T_2$ output, or both. The values are the average of results from four datasets.}
    \label{tab:temporal_target_lscd}
    \footnotesize
    \setlength{\tabcolsep}{1.5pt}
    \renewcommand{\arraystretch}{0.95}
    \begin{tabular*}{\linewidth}{@{\extracolsep{\fill}}l|c|ccc|ccc|ccc}
        \toprule
        \multirow{2}{*}{Model}
        & \multirow{2}{*}{Fusion}
        & \multicolumn{3}{c|}{Target $\mathcal{L}_{\mathrm{scd}}^{T_{1}}$}
        & \multicolumn{3}{c|}{Target $\mathcal{L}_{\mathrm{scd}}^{T_{2}}$}
        & \multicolumn{3}{c}{Target $\mathcal{L}_{\mathrm{scd}}$} \\
        \cmidrule(lr){3-5}
        \cmidrule(lr){6-8}
        \cmidrule(lr){9-11}
        &
        & $\mathrm{ASR}_{\mathrm{scd}}^{T_1}$
        & $\mathrm{ASR}_{\mathrm{scd}}^{T_2}$
        & $\mathrm{ASR}_{\mathrm{scd}}$
        & $\mathrm{ASR}_{\mathrm{scd}}^{T_1}$
        & $\mathrm{ASR}_{\mathrm{scd}}^{T_2}$
        & $\mathrm{ASR}_{\mathrm{scd}}$
        & $\mathrm{ASR}_{\mathrm{scd}}^{T_1}$
        & $\mathrm{ASR}_{\mathrm{scd}}^{T_2}$
        & $\mathrm{ASR}_{\mathrm{scd}}$ \\
        \midrule
        HRSCD4      & Concat.           & 90.16 & 41.82 & 66.51 & 43.64 & 89.84 & 66.22 & 86.28 & 84.51 & 85.36 \\
        EGMSNet     & Diff.             & 91.51 & 55.20 & 73.56 & 57.41 & 89.37 & 73.15 & 90.96 & 88.49 & 89.62 \\
        BGSNet      & Cross-Attn.+Diff. & 92.94 & 70.02 & 81.82 & 70.92 & 93.95 & 82.05 & 89.12 & 89.47 & 89.20 \\
        CPGNet      & Diff.+Concat.     & 95.27 & 71.39 & 83.44 & 69.05 & 94.75 & 81.65 & 88.89 & 90.48 & 89.59 \\
        ChangeMamba & SSM+Concat.       & 95.57 & 70.39 & 83.07 & 69.64 & 95.02 & 82.25 & 90.05 & 89.33 & 89.69 \\
        FoBa        & Mul.+Concat.      & 94.83 & 68.29 & 81.53 & 68.44 & 93.51 & 80.99 & 86.58 & 83.81 & 85.20 \\
        \bottomrule
    \end{tabular*}
\end{table*}

The semantic+binary-change and complete joint objectives also remain below the semantic-change objective in their averaged $\mathrm{ASR}_{\mathrm{scd}}$ ranges (67.01--80.74\% and 77.78--84.74\%, respectively). FoBa is the sole exception to the overall objective ordering: its binary-change attack reaches 87.47\%, compared with 85.20\% for the semantic-change attack. Supplementary Fig.~S1 illustrates the associated qualitative failure modes. We retain the semantic-change objective because it is the strongest final-output target for most model averages.

\subsubsection{Temporal Semantic-Change Targets}
\label{subsubsec:temporal_target_results}
We next examine whether adversarial effects remain localized to the targeted temporal output. Perturbation access is fixed to the joint bitemporal setting, allowing both $T_1$ and $T_2$ inputs to be modified. We vary only the output-side objective by optimizing the $T_1$ semantic-change loss $\mathcal{L}_{\mathrm{scd}}^{T_1}$, the $T_2$ semantic-change loss $\mathcal{L}_{\mathrm{scd}}^{T_2}$, or their sum $\mathcal{L}_{\mathrm{scd}}=\mathcal{L}_{\mathrm{scd}}^{T_1}+\mathcal{L}_{\mathrm{scd}}^{T_2}$.

Table~\ref{tab:temporal_target_lscd} reports model-level averages; Supplementary Table~S4 provides the dataset--model breakdown. A single-side objective affects its corresponding output most strongly for every model. The $T_1$ target yields 90.16--95.57\% on the targeted side and 41.82--71.39\% on the opposite side; the $T_2$ target yields corresponding ranges of 89.37--95.02\% and 43.64--70.92\%. Thus, temporal effects are localized but not isolated.

The joint objective produces more balanced timestamp-specific averages (86.28--90.96\% for $T_1$ and 83.81--90.48\% for $T_2$) and a higher overall ASR than either single-side objective for every model in Table~\ref{tab:temporal_target_lscd}.

Supplementary Fig.~S2 provides a spatial example of the same pattern: single-side objectives concentrate errors on the targeted prediction, whereas joint optimization distributes degradation more evenly across both outputs.



\begin{table*}[t]
    \centering
    \caption{Attack success rates for PGD access to $T_1$, $T_2$, or both inputs, using the semantic-change target. The values are the average of results from four datasets.}
    \label{tab:cross_temporal_model_extended}
    \scriptsize
    \setlength{\tabcolsep}{1.5pt}
    \renewcommand{\arraystretch}{0.95}
    \begin{tabular*}{\linewidth}{@{\extracolsep{\fill}}lc|ccc|ccc|ccc@{}}
        \toprule
        \multirow{2}{*}{Model}
        & \multirow{2}{*}{Fusion}
        & \multicolumn{3}{c|}{$T_1$ Perturbed}
        & \multicolumn{3}{c|}{$T_2$ Perturbed}
        & \multicolumn{3}{c}{$T_1+T_2$ Perturbed} \\
        \cmidrule(lr){3-5}
        \cmidrule(lr){6-8}
        \cmidrule(lr){9-11}
        &
        & $\mathrm{ASR}_{\mathrm{scd}}^{T_1}$
        & $\mathrm{ASR}_{\mathrm{scd}}^{T_2}$
        & $\mathrm{ASR}_{\mathrm{scd}}$
        & $\mathrm{ASR}_{\mathrm{scd}}^{T_1}$
        & $\mathrm{ASR}_{\mathrm{scd}}^{T_2}$
        & $\mathrm{ASR}_{\mathrm{scd}}$
        & $\mathrm{ASR}_{\mathrm{scd}}^{T_1}$
        & $\mathrm{ASR}_{\mathrm{scd}}^{T_2}$
        & $\mathrm{ASR}_{\mathrm{scd}}$ \\
        \midrule
        HRSCD4      & Concat.           & 85.15 & 36.38 & 61.12 & 34.53 & 82.94 & 58.20 & 86.25 & 84.57 & 85.36 \\
        EGMSNet     & Diff.             & 88.26 & 29.95 & 59.66 & 24.89 & 82.30 & 53.04 & 90.92 & 88.40 & 89.62 \\
        BGSNet      & Cross-Attn.+Diff. & 83.95 & 52.20 & 68.45 & 48.99 & 88.17 & 68.02 & 89.13 & 89.35 & 89.20 \\
        CPGNet      & Diff.+Concat.     & 90.61 & 45.83 & 68.46 & 40.95 & 89.12 & 64.61 & 88.97 & 90.25 & 89.59 \\
        ChangeMamba & SSM+Concat.       & 90.19 & 31.86 & 61.23 & 27.35 & 89.39 & 58.05 & 90.11 & 89.29 & 89.69 \\
        FoBa        & Mul.+Concat.      & 86.83 & 59.18 & 72.93 & 57.81 & 82.25 & 70.04 & 86.61 & 83.87 & 85.20 \\
        \bottomrule
    \end{tabular*}
\end{table*}

\subsection{Temporal Perturbation Access Analysis}
\label{subsec:temporal_access_results}

We next examine how restricting adversarial access to different temporal inputs affects SCD robustness. The attack objective is fixed to the semantic-change loss $\mathcal{L}_{\mathrm{scd}}$, while perturbations are applied to $T_1$, $T_2$, or both inputs. This experiment therefore isolates input-side temporal vulnerability under a common task-level objective. Table~\ref{tab:cross_temporal_model_extended} reports the model-level timestamp-specific and overall attack success rates, Supplementary Table~S5 provides the dataset--model breakdown, and Supplementary Fig.~S3 illustrates the resulting spatial errors.

Single-temporal attacks show an access-aligned pattern: perturbing $T_1$ causes greater degradation on the $T_1$ output, and perturbing $T_2$ causes greater degradation on the $T_2$ output. This diagonal tendency is also visible in Supplementary Fig.~S3.
The effects are not confined to the directly corresponding output: opposite-side averages are 29.95--59.18\% under $T_1$-only access and 24.89--57.81\% under $T_2$-only access. Thus, an unmodified image does not imply an unchanged prediction for its timestamp.

\begin{table}[t]
\centering
\caption{$\mathrm{ASR}_{\mathrm{scd}}$ (\%) for perturbation methods. All methods are applied to both temporal inputs; gradient-based attacks use the semantic-change target. The values are the average of results from four datasets.}
\label{tab:attack_method_asr_with_pgd}
\resizebox{\linewidth}{!}{
\begin{tabular}{l|cccccc}
\toprule
 Model
& Gaussian
& Jitter
& FGSM
& I-FGSM
& C\&W
& PGD \\
\midrule
HRSCD4      & 0.02 & 1.41 & 48.61 & 86.46 & 74.47 & 85.36 \\
EGMSNet     & 0.02 & 3.16 & 59.78 & 90.66 & 72.68 & 89.62 \\
BGSNet      & 0.02 & 1.95 & 49.80 & 89.75 & 76.10 & 89.20 \\
CPGNet      & 0.05 & 2.38 & 48.55 & 90.95 & 73.01 & 89.59 \\
ChangeMamba & 0.05 & 2.07 & 47.43 & 90.95 & 67.46 & 89.69 \\
FoBa        & 0.02 & 1.55 & 49.55 & 85.92 & 68.52 & 85.20 \\
\bottomrule
\end{tabular}
}
\end{table}

Joint access produces the largest overall degradation for all six model averages: 85.20--89.69\%, compared with 59.66--72.93\% for $T_1$-only access and 53.04--70.04\% for $T_2$-only access. It also exposes more input dimensions because both timestamps receive separately constrained perturbations. Section~\ref{sec:discussion} discusses the broader implications for temporal locality and coupling.

\begin{table*}[t]
    \centering
    \caption{Cross-architecture PGD transferability with the semantic-change target. Rows are source models, columns are target models, and entries are $\mathrm{ASR}_{\mathrm{scd}}$ values. The values are the average of results from four datasets. Diagonal entries are white-box attacks noted in \textbf{bold}.}
    \label{tab:cross_architecture_transfer_datasets}
    \footnotesize
    \setlength{\tabcolsep}{2.5pt}
    \renewcommand{\arraystretch}{0.95}
    \begin{tabular*}{\linewidth}{@{\extracolsep{\fill}}ll|cccccc@{}}
        \toprule
        \multirow{2}{*}{Source}
        & \multirow{2}{*}{Family}
        & \multicolumn{6}{c}{Target model} \\
        &
        & HRSCD4
        & EGMSNet
        & BGSNet
        & CPGNet
        & ChangeMamba
        & FoBa \\
        \midrule
        HRSCD4      & CNN         & \textbf{85.36} & 0.62 & 0.53 & 0.80 & 0.73 & 0.59 \\
        EGMSNet     & CNN         & 1.05 & \textbf{89.62} & 2.72 & 3.47 & 3.64 & 3.35 \\
        BGSNet      & Transformer & 0.59 & 1.94 & \textbf{89.20} & 12.76 & 3.55 & 3.05 \\
        CPGNet      & Transformer & 0.87 & 3.42 & 26.59 & \textbf{89.59} & 7.29 & 6.18 \\
        ChangeMamba & Mamba       & 0.76 & 2.92 & 4.72 & 5.82 & \textbf{89.69} & 10.34 \\
        FoBa        & Mamba       & 0.84 & 2.76 & 4.50 & 5.07 & 11.54 & \textbf{85.20} \\
        \bottomrule
    \end{tabular*}
\end{table*}

\begin{table*}[!htb]
\centering
\caption{Error-mode rates (\%) under default joint bitemporal PGD with the semantic-change target. Columns report attack success on $F_{\mathrm{scd}}$, change suppression, false-change hallucination, and transition corruption.}
\label{tab:scd_error_patterns}
\footnotesize
\setlength{\tabcolsep}{1.2pt}
\renewcommand{\arraystretch}{0.95}
\begin{tabular*}{\linewidth}{@{\extracolsep{\fill}}l|cccc|cccc|cccc|cccc@{}}
\toprule
\multirow{2}{*}{Model}
& \multicolumn{4}{c|}{HiUCD-Mini}
& \multicolumn{4}{c|}{JL1}
& \multicolumn{4}{c|}{LandsatSCD}
& \multicolumn{4}{c}{SECOND} \\
\cmidrule(lr){2-5}
\cmidrule(lr){6-9}
\cmidrule(lr){10-13}
\cmidrule(lr){14-17}
& ASR & $R_{\mathrm{sup}}$ & $R_{\mathrm{hal}}$ & $R_{\mathrm{trans}}$
& ASR & $R_{\mathrm{sup}}$ & $R_{\mathrm{hal}}$ & $R_{\mathrm{trans}}$
& ASR & $R_{\mathrm{sup}}$ & $R_{\mathrm{hal}}$ & $R_{\mathrm{trans}}$
& ASR & $R_{\mathrm{sup}}$ & $R_{\mathrm{hal}}$ & $R_{\mathrm{trans}}$ \\
\midrule
HRSCD4      & 97.97 & 10.13 & 12.04 & 99.96 & 99.08 & 5.68 & 67.44 & 99.96 & 55.53 & 33.17 & 3.90 & 70.31 & 88.87 & 32.04 & 28.72 & 98.85 \\
EGMSNet     & 96.68 & 2.54 & 10.39 & 99.69 & 97.34 & 1.82 & 44.11 & 99.98 & 75.18 & 26.11 & 14.31 & 75.15 & 89.29 & 6.44 & 24.09 & 98.71 \\
BGSNet      & 92.78 & 2.35 & 17.57 & 99.26 & 89.19 & 34.38 & 18.76 & 98.44 & 83.42 & 20.90 & 23.53 & 84.02 & 91.42 & 3.46 & 47.14 & 97.76 \\
CPGNet      & 89.62 & 7.88 & 7.91 & 98.69 & 86.29 & 33.01 & 18.07 & 98.09 & 87.66 & 34.82 & 10.09 & 89.60 & 94.77 & 17.56 & 33.64 & 98.85 \\
ChangeMamba & 94.56 & 9.30 & 7.28 & 99.41 & 88.88 & 40.48 & 20.12 & 98.53 & 83.69 & 18.35 & 17.94 & 90.68 & 91.62 & 5.54 & 36.93 & 98.12 \\
FoBa        & 91.77 & 18.55 & 5.19 & 99.17 & 79.79 & 44.33 & 20.39 & 96.71 & 87.81 & 16.22 & 20.09 & 90.09 & 81.43 & 11.70 & 25.14 & 95.45 \\
\bottomrule
\end{tabular*}
\end{table*}

\subsection{Attack Generation Method Comparison}
\label{subsec:attack_method_results}

We examine whether the observed vulnerability depends on the perturbation generator. Under joint bitemporal access, we compare Gaussian noise and jitter with FGSM, I-FGSM, PGD, and the C\&W-style margin attack introduced in Section~\ref{subsec:attack_generation}. These methods cover non-optimized disturbances, single-step gradient optimization, and iterative adversarial optimization.

Table~\ref{tab:attack_method_asr_with_pgd} shows the model-level averages, while Supplementary Table~S6 provides the dataset--model results. Both presentations show a consistent separation between the non-optimized and gradient-based methods. Gaussian noise and jitter cause limited degradation under the evaluated settings, whereas all four gradient-based attacks are more effective.

FGSM is less effective and more variable than I-FGSM and PGD. The two iterative gradient-sign methods produce similarly high degradation across the evaluated model families. Supplementary Fig.~S4 shows examples ranging from disappearance of predicted regions to semantic substitution and widespread output corruption.

The C\&W-style configuration is intermediate and more dataset dependent. Its results remain below those of I-FGSM and PGD despite iterative optimization. Because I-FGSM and PGD provide comparable attack strength, PGD is retained as the default generator for consistency with the preceding experiments.


\begin{figure*}[!htb]
    \centering
    \includegraphics[width=.97\linewidth]{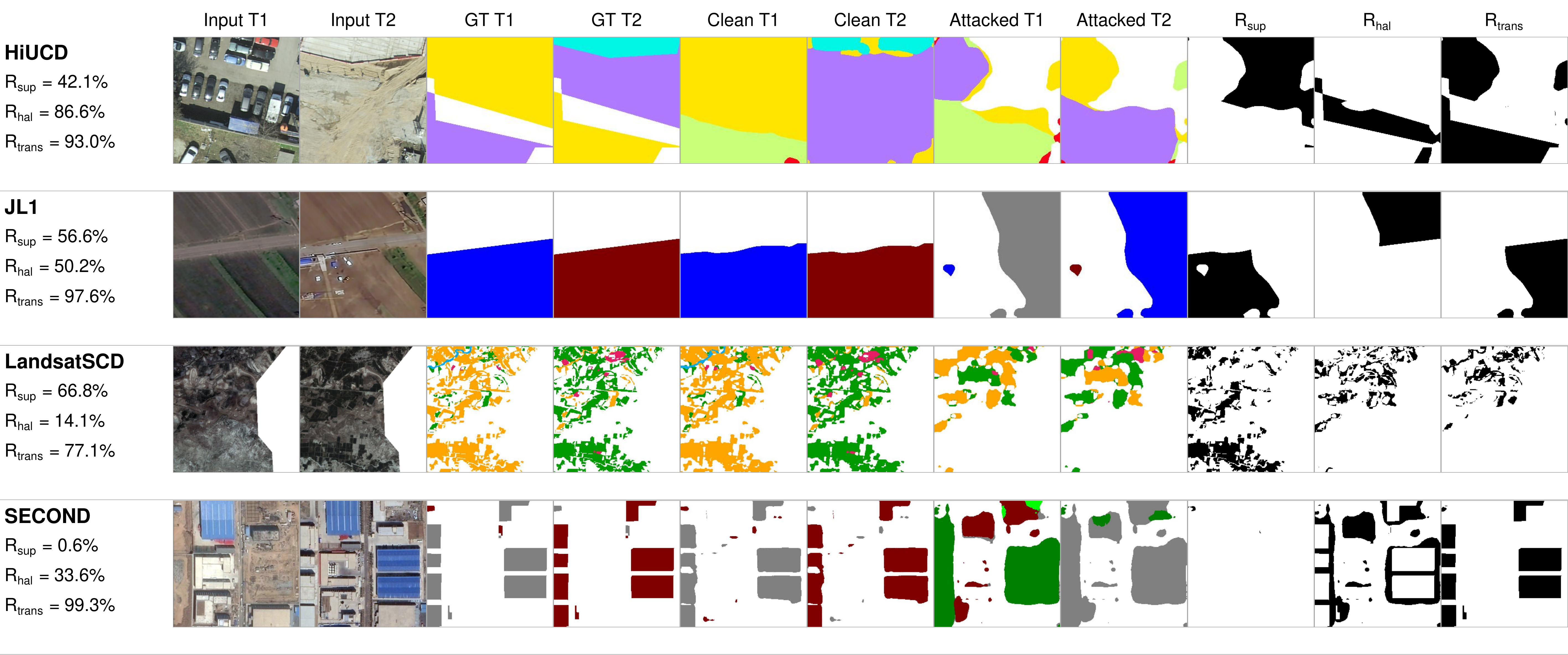}
    \caption{SCD error modes for BGSNet under default joint bitemporal PGD with the semantic-change target. Rows are datasets; columns show inputs, ground truth, clean and adversarial predictions, and masks for suppression ($R_{\mathrm{sup}}$), hallucination ($R_{\mathrm{hal}}$), and transition corruption ($R_{\mathrm{trans}}$). Black marks affected pixels; percentages follow Eq.~\eqref{eq:failure_rates_compact}.}
    \label{fig:qualitative_error_modes}
\end{figure*}

\subsection{Cross-Architecture Transferability}
\label{subsec:transferability_results}

We further investigate whether adversarial perturbations generated on one SCD model transfer to other architectures. This represents a transfer-based black-box setting in which a surrogate source model is used to generate perturbations that are subsequently evaluated on a different target model. PGD is applied jointly to both temporal inputs using the semantic-change objective. Rows in Table~\ref{tab:cross_architecture_transfer_datasets} denote source models, while columns denote target models; Supplementary Table~S7 provides the corresponding dataset-specific matrices.

The averaged transfer matrix shows a pronounced gap between the white-box diagonal and off-diagonal attacks. Diagonal $\mathrm{ASR}_{\mathrm{scd}}$ values are 85.20--89.69\%, with a mean of 88.11\%, whereas the off-diagonal mean is only 4.43\%. The largest off-diagonal value is 26.59\% for CPGNet-to-BGSNet transfer, compared with 12.76\% in the reverse direction. Under this direct-transfer protocol, most perturbations therefore lose the majority of their effectiveness when evaluated on a different model.

The largest off-diagonal entries often occur between models with related encoders. BGSNet and CPGNet, which share a PVTv2-B2 encoder, transfer more strongly to each other than to most unrelated models. A weaker corresponding pattern appears between ChangeMamba and FoBa, whereas transfer between the two CNN models remains limited. These are associations within the evaluated model set; representation and gradient alignment are not measured. The off-diagonal matrix is also asymmetric: models that are effective sources are not necessarily the most susceptible targets. Section~\ref{sec:discussion} considers the implications and scope of this direct-transfer result.

\subsection{SCD-Specific Failure Diagnosis}
\label{subsec:failure_diagnosis}


\subsubsection{Error-Mode Analysis}
\label{subsubsec:error_mode_results}

Fig.~\ref{fig:qualitative_error_modes} illustrates change suppression, false-change hallucination, and semantic-transition corruption, and Table~\ref{tab:scd_error_patterns} reports their diagnostic rates. The same numerical results are reproduced in Supplementary Table~S8 for completeness. Because the rates use different denominators, they describe separate conditional behaviors rather than comparable shares of the total error.

Conditional transition-corruption rates are high on HiUCD-Mini, JL1, and SECOND: most ground-truth changes that remain detected after attack are assigned an incorrect transition. This conditional rate does not measure the absolute number of corrupted pixels.

The two localization-related rates vary more strongly across datasets and models. HiUCD-Mini combines comparatively limited suppression and hallucination with high transition corruption. JL1 has a mixed profile in which suppression or hallucination dominates for different models, while SECOND combines high transition corruption with substantial false-change responses.

LandsatSCD has lower transition-corruption rates than the other datasets and more pronounced suppression for several models. The diagnostic profile therefore differs even where aggregate $\mathrm{ASR}_{\mathrm{scd}}$ values alone appear similar.

\subsubsection{Class-Wise Vulnerability}
\label{subsubsec:classwise_vulnerability}

Supplementary Table~S9 provides the class-wise results together with class frequency and clean class-wise performance. These quantities must be interpreted jointly: a high class-wise ASR is informative only when the class has sufficient support and a meaningful clean $F_{\mathrm{scd}}^{c}$, while a low ASR for a rare class or a class with negligible clean performance does not establish robustness. The supplementary results show that class-wise vulnerability varies across datasets and models, so no single semantic category or architecture admits a universal robustness ordering.



\section{Discussion}
\label{sec:discussion}


The experiments show that adversarial robustness in SCD is shaped by the interaction among output coupling, temporal information flow, and the model prediction pathway. The most consequential observation is that robustness measured on an individual component does not necessarily represent robustness of the final semantic-change prediction. The temporal experiments further show that prediction errors are not confined to the directly targeted output or perturbed input. These findings make SCD a distinct robustness problem from conventional single-input dense prediction.

\subsection{Task-Aligned Vulnerability across Outputs}
\label{subsec:discussion_coupled}

The central result is that SCD robustness cannot be inferred from binary change localization or timestamp-wise semantic recognition in isolation. A model can retain much of the changed/unchanged structure while assigning the wrong land-cover states or transition direction. This mismatch makes the final semantic-change output the task-defining object of robustness assessment: a spatially plausible change mask is not a reliable proxy for a correct final prediction.

The attack-objective comparison explains this mismatch. Binary-change attacks primarily disturb localization, semantic-branch attacks alter the land-cover evidence at each timestamp, and semantic-change attacks directly optimize the output used for task-level assessment. Direct semantic-change optimization is the most consistent objective overall, although the degree of separation among objectives remains dataset dependent. Moreover, adding binary and semantic losses does not consistently strengthen the attack. The supported interpretation is therefore not that one branch is universally dominant, but that attack strength depends more on alignment with the evaluated output than on the number of activated loss components.

This result shifts the unit of robustness analysis from an individual branch to the complete SCD decision pathway. Coupling semantic recognition, change localization, and final decoding is necessary for the task, but it also permits errors to move between components. The experiments do not establish that tighter coupling is inherently less robust; such a causal claim would require controlled ablations of fusion, gating, supervision, and decoding. They do show that a defense protecting only the binary mask or one semantic branch cannot be assumed to protect the final result. Robust SCD requires both branch-level stability and consistency at the interfaces where semantic and change evidence are combined.

\subsection{Cross-Temporal Propagation and Model Design}
\label{subsec:discussion_temporal}

The temporal experiments expose two distinct forms of dependency. Temporal output targeting changes which timestamp-wise prediction is optimized while both inputs remain accessible, whereas temporal perturbation access changes which input can be modified while the final output objective remains fixed. The former diagnoses propagation between output branches; the latter diagnoses how an accessible observation influences the bitemporal prediction pathway. Keeping these settings separate is important because they represent different attacker capabilities and answer different robustness questions.

Both experiments nevertheless reveal the same higher-level pattern: temporal locality coexists with cross-temporal propagation. An attack has its greatest effect on the directly targeted or perturbed side, yet it also affects the prediction associated with the other timestamp. An unmodified observation therefore does not guarantee a stable prediction for its timestamp. Joint input access exposes a broader attack surface and should be interpreted as a more permissive threat model rather than as a direct efficiency comparison with single-timestamp access.

Cross-temporal effects are approximately bidirectional at the aggregate level, although their magnitude varies across datasets and models. These variations do not produce a consistent ordering of CNN-, Transformer-, and state-space-based architectures, and stronger clean performance does not reliably imply greater adversarial stability. Broad encoder labels are consequently less informative than the complete prediction pathway, including representation, temporal interaction, fusion, supervision, and decoding. Future models should make temporal interaction more stable and inspectable through mechanisms such as robust alignment, branch-specific uncertainty, perturbation-stable fusion, and consistency checks between timestamp-wise semantics and the change mask. Such mechanisms should be evaluated under both single- and joint-access settings, because aggregate $F_{\mathrm{scd}}$ alone cannot distinguish direct sensitivity from propagated error.

\subsection{Threat Models and Failure-Aware Evaluation}
\label{subsec:discussion_protocol}

The attack-method comparison demonstrates that robustness claims are conditional on how the stress test is constructed. Optimized iterative attacks reveal failures that are not captured by simple non-optimized disturbances. This comparison distinguishes adversarial vulnerability from the evaluated baseline disturbances, but it neither establishes robustness to natural corruption nor exhausts the space of possible attacks.

Direct cross-architecture transfer is substantially weaker than white-box attack, indicating that many effective perturbation directions are model specific under the evaluated direct-transfer protocol. Transfer nevertheless remains dependent on the source, target, and data context. Weak average transfer should therefore not be interpreted as black-box security: stronger surrogate construction, transfer-enhancing objectives, target-model queries, and substitute-model adaptation were not evaluated.

A credible SCD robustness claim must consequently specify the optimized output, accessible timestamps, perturbation constraint, attack generator, initialization, optimization settings, preprocessing, valid-pixel handling, and decoding procedure. It should report component-, timestamp-, and final-output performance under a clearly stated threat model. This reporting is not merely procedural detail: two evaluations using the same perturbation budget can represent materially different risks. Protocol transparency makes robustness results more reproducible, auditable, and accountable, and reduces the chance that objective mismatch or restricted access is mistaken for stable model behavior.

Aggregate attack success is also insufficient because it measures the magnitude, not the form, of failure. Change suppression can conceal genuine transitions, false-change hallucination can generate unnecessary alarms, and semantic-transition corruption can preserve plausible spatial support while misrepresenting what changed. Conditional transition corruption is the most recurrent diagnostic pattern in the present experiments, while suppression, hallucination, and class-wise vulnerability vary more strongly across datasets and models. These rates use different denominators and are not additive; class-wise results can also be misleading when clean performance or support is low. Robustness reports should therefore pair aggregate scores with absolute or conditional error counts, clean class-wise performance, and transition support. For trustworthy AI, the value of this diagnosis is not simply to rank models, but to make visible how an apparently credible Earth-observation product can become semantically unreliable \cite{xu2023ai,ghamisi2025responsible}.

\subsection{Scope and Limitations}
\label{subsec:discussion_limitations}

The primary limitations concern security scope, attribution, and statistical stability. The evaluated digital, norm-bounded threat model provides a controlled high-access stress test but does not represent physical acquisition effects, operational transformations, natural corruption, adaptive black-box attacks, targeted manipulation, or training-time threats such as data poisoning and backdoors. The architecture and dataset comparisons are also observational: models and datasets differ along multiple dimensions, so the reported patterns establish associations rather than causes and remain limited to paired-image SCD and the evaluated model families. In addition, relative degradation, rare-class estimates, and conditional failure rates require careful interpretation when clean performance or support is limited. Future work should combine sensor-aware and operational attacks with controlled ablations, matched data conditions, repeated trials, uncertainty estimates, absolute error counts, and transition-level support \cite{hao2025towards}.

This study evaluates vulnerabilities rather than mitigation strategies. A defense for SCD should be tested against attacks on the semantic branches, binary change localization, and final semantic-change outputs, as well as under different temporal-access settings. Improving one component may not prevent errors from propagating between timestamps or appearing during final-output decoding. This issue may become more important as SCD systems incorporate geospatial foundation models, vision-language components, or generative modules, although these model classes are outside the scope of the present experiments.

Adversarial robustness alone does not establish that an AI system is ready for responsible deployment. Data quality, uncertainty, privacy, fairness, and human oversight also require separate assessment \cite{ghamisi2025responsible,suhong2026operational}. For SCD, this distinction has practical relevance because concealed changes, false alarms, and incorrect transitions can lead to different errors in environmental monitoring and planning. The proposed framework should therefore be viewed as a task-specific security test rather than a complete assurance framework for AI deployment.

\section{Conclusion}
\label{sec:conclusion}

This work studied adversarial robustness in semantic change detection as a coupled bitemporal prediction problem. By separating output-side attack targets from input-side temporal access, the proposed framework distinguishes vulnerability of individual prediction components from propagation through temporal information pathways.

The experiments reveal three main behaviors. First, the robustness of binary change localization or an individual semantic branch does not reliably represent the robustness of the final semantic-change prediction. Second, adversarial effects are temporally local but not temporally isolated: perturbing or targeting one side of the bitemporal system can alter predictions associated with the other. Third, these vulnerabilities occur across different architecture families, while clean performance and broad backbone type provide limited indication of adversarial stability.

These findings suggest that robustness in bitemporal dense prediction should be evaluated at the level of the complete decision pathway. Future work should investigate defenses that explicitly stabilize temporal interaction and coupled output decoding, as well as more general attack settings and multi-input vision tasks. The proposed framework provides a basis for analyzing these dependencies in SCD and related bitemporal image-understanding systems.

\section*{Acknowledgment}
Weikang Yu's work was supported by the 3D-ABC project as part of the Helmholtz Foundation Model Initiative. The authors would also like to thank the European Regional Development Fund and the Land of Saxony for providing the high-specification Nvidia A100 GPU servers that we used in our experiments.

\bibliographystyle{unsrt}
\bibliography{reference}

\clearpage
\onecolumn

\setcounter{section}{0}
\setcounter{subsection}{0}
\setcounter{subsubsection}{0}
\setcounter{figure}{0}
\setcounter{table}{0}
\setcounter{equation}{0}
\renewcommand{\thesection}{S\arabic{section}}
\renewcommand{\thesubsection}{\thesection.\Alph{subsection}}
\renewcommand{\thesubsubsection}{\thesubsection.\arabic{subsubsection}}
\renewcommand{\thetable}{S\arabic{table}}
\renewcommand{\thefigure}{S\arabic{figure}}
\renewcommand{\theequation}{S\arabic{equation}}
\renewcommand{\theHsection}{supp.\arabic{section}}
\renewcommand{\theHsubsection}{supp.\arabic{section}.\arabic{subsection}}
\renewcommand{\theHsubsubsection}{supp.\arabic{section}.\arabic{subsection}.\arabic{subsubsection}}
\renewcommand{\theHtable}{supp.\arabic{table}}
\renewcommand{\theHfigure}{supp.\arabic{figure}}
\renewcommand{\theHequation}{supp.\arabic{equation}}
\renewcommand{\topfraction}{0.95}
\renewcommand{\bottomfraction}{0.90}
\renewcommand{\textfraction}{0.05}
\renewcommand{\floatpagefraction}{0.85}
\setlength{\textfloatsep}{12pt plus 2pt minus 2pt}
\setlength{\floatsep}{12pt plus 2pt minus 2pt}
\setlength{\intextsep}{12pt plus 2pt minus 2pt}
\makeatletter
\setlength{\@fptop}{0pt}
\setlength{\@fpbot}{0pt plus 1fil}
\makeatother

\begin{center}
{\LARGE\itshape Supplementary Materials for\par}
\vspace{0.45em}
{\LARGE On the Adversarial Robustness of Remote Sensing Semantic Change Detection\par}
\end{center}
\vspace{1.25em}
\section{Additional Qualitative Results}
\begin{figure}[!htbp]
    \centering
    \includegraphics[width=.96\linewidth]{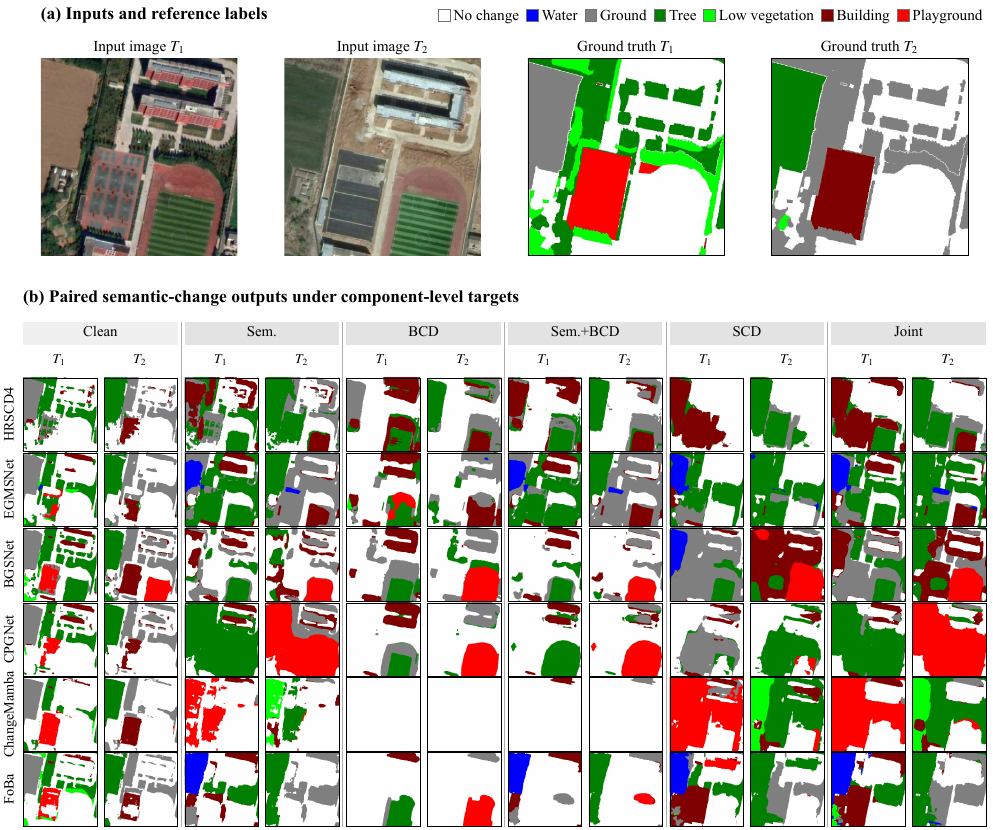}
    \caption{Component-level attack objectives on one SECOND sample under joint bitemporal perturbation. Paired columns show the final $T_1$ and $T_2$ semantic-change predictions. The examples illustrate distinct failure modes, including semantic replacement, change suppression, change-region expansion, and widespread corruption. These observations illustrate individual model behavior on the selected sample and are not intended as general architectural conclusions.}
    \label{supp:fig:component_target_qualitative}
\end{figure}

\begin{figure}[!htbp]
    \centering
    \includegraphics[width=.92\linewidth]{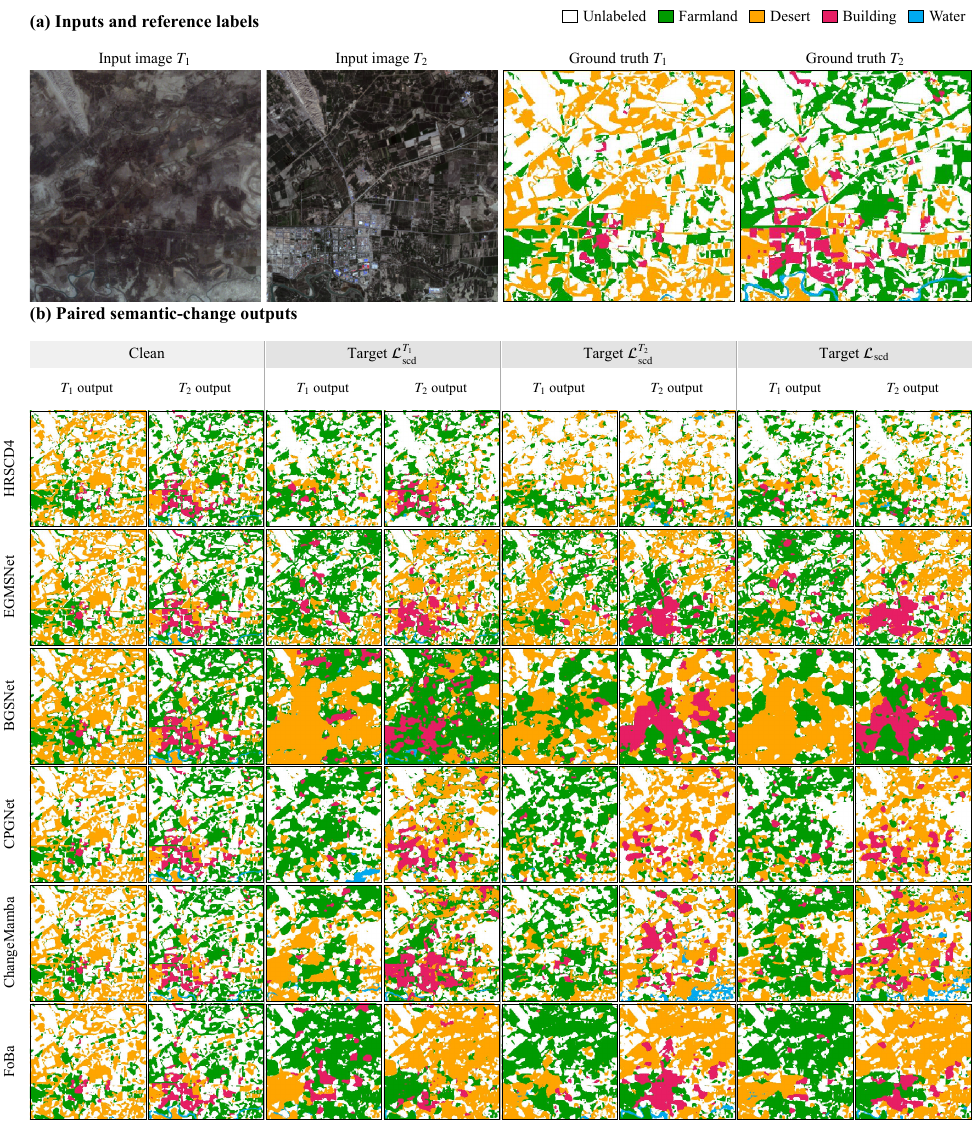}
    \caption{Temporal semantic-change targets on one LandsatSCD sample. Both inputs are perturbed while the loss targets the $T_1$ output, the $T_2$ output, or both.}
    \label{supp:fig:temporal_target_qualitative}
\end{figure}

\begin{figure}[!htbp]
    \centering
    \includegraphics[width=.92\linewidth]{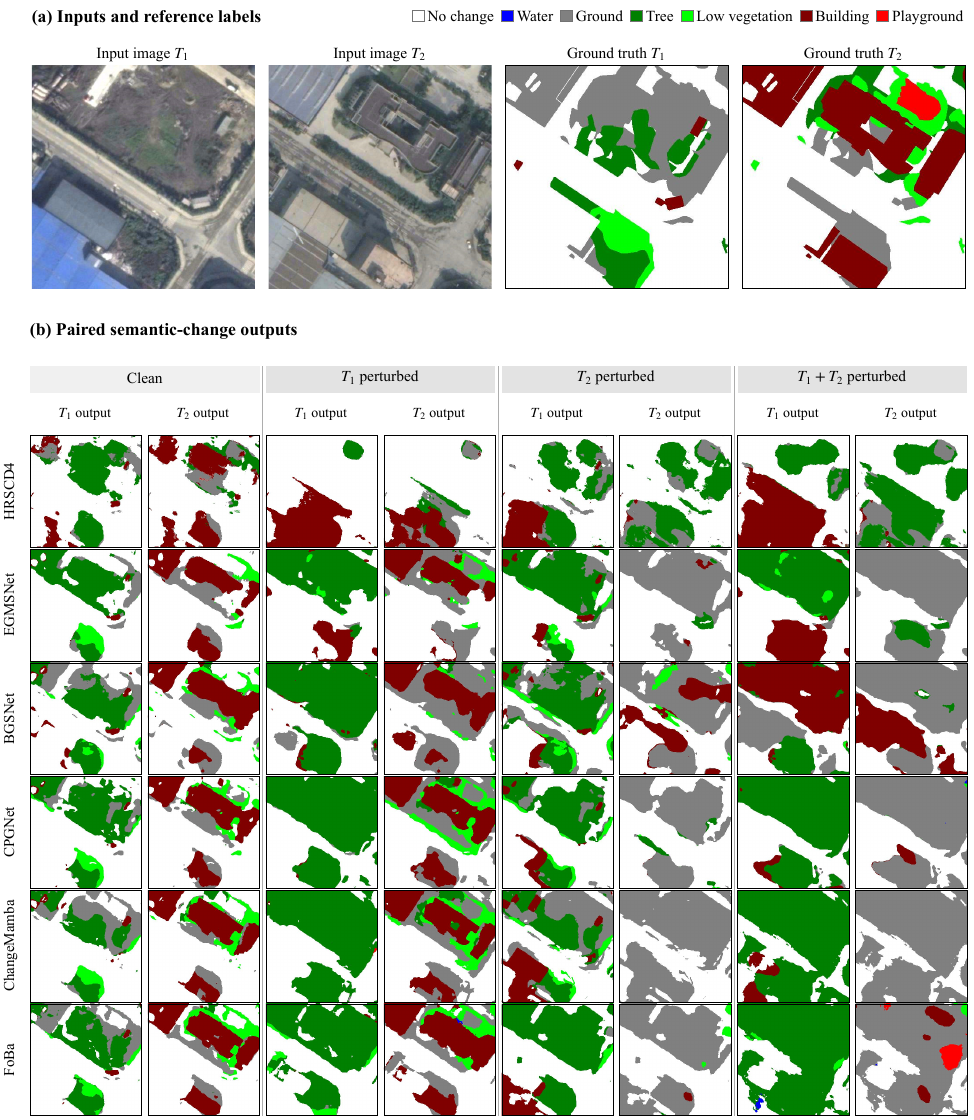}
    \caption{Temporal perturbation access on one SECOND sample. Each model row shows paired $T_1$ and $T_2$ predictions for clean, $T_1$-only, $T_2$-only, and joint perturbations. Opposite-side changes indicate cross-temporal propagation.}
    \label{supp:fig:temporal_access_qualitative}
\end{figure}

\begin{figure}[!htbp]
    \centering
    \includegraphics[width=.92\linewidth]{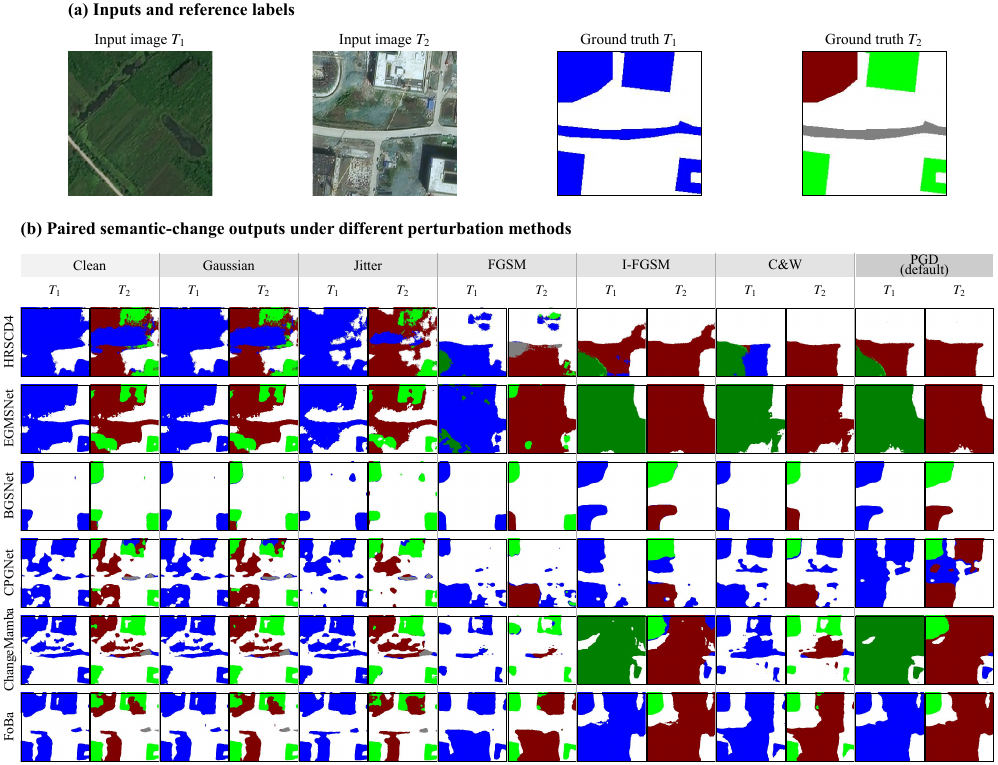}
    \caption{Perturbation methods on one JL1 sample. Gaussian noise and jitter largely preserve the clean outputs, whereas gradient-based attacks cause semantic corruption; PGD is the default.}
    \label{supp:fig:attack_method_qualitative}
\end{figure}

We provide additional qualitative semantic change detection visualization results in Figures~\ref{supp:fig:component_target_qualitative}--\ref{supp:fig:attack_method_qualitative}. The results are shown for the six evaluated models on representative samples from the JL1, SECOND, and LandsatSCD datasets under different perturbation methods, attack objectives, temporal targets, and temporal access settings. Specifically, Figure~\ref{supp:fig:component_target_qualitative} presents component-level attack objectives on SECOND; Figure~\ref{supp:fig:temporal_target_qualitative} examines temporal semantic-change targets on LandsatSCD; and Figure~\ref{supp:fig:temporal_access_qualitative} shows the effects of perturbing $T_1$, $T_2$, or both inputs on SECOND;
Figure~\ref{supp:fig:attack_method_qualitative} compares Gaussian noise, jitter, FGSM, I-FGSM, C\&W, and PGD on JL1; 
It can be observed that Gaussian noise and jitter largely preserve the clean outputs, whereas gradient-based attacks cause substantial semantic corruption. Moreover, the component and temporal results reveal diverse failure modes, including semantic replacement, change suppression, change-region expansion, widespread corruption, and cross-temporal propagation.

\section{Detailed Comparison Results}
We provide additional quantitative comparison results in Tables~\ref{supp:tab:clean_performance}--\ref{supp:tab:class_wise_vulnerability}. Unless otherwise stated, all attack success rates are relative reductions from the corresponding clean score, and all attacks follow the conventions defined in the main paper. Specifically, Table~\ref{supp:tab:clean_performance} reports the clean performance of the evaluated models, while Table~\ref{supp:tab:standard_pgd_robustness} presents their robustness under the default white-box joint bitemporal PGD setting. Tables~\ref{supp:tab:objective_comparison_diagnostic}--\ref{supp:tab:cross_temporal_model_extended} examine component-level objectives, temporal targets, and temporal access patterns. Table~\ref{supp:tab:attack_method_asr_with_pgd} compares different perturbation methods, while Tables~\ref{supp:tab:cross_architecture_transfer_datasets}--\ref{supp:tab:class_wise_vulnerability} analyze cross-architecture transferability, error modes, and class-wise vulnerability, respectively. It can be observed that Gaussian noise and jitter cause limited performance degradation, whereas gradient-based attacks substantially reduce semantic change detection performance. Moreover, white-box attacks are generally much stronger than cross-architecture transferred attacks, and the diagnostic results show that attack behavior varies with the objective, temporal access pattern, dataset, model, and semantic class.

\FloatBarrier
\clearpage

\begin{table}[!htbp]
    \centering
    \caption{Clean performance of the evaluated models on four SCD datasets (\%).}
    \label{supp:tab:clean_performance}
    \resizebox{.5\linewidth}{!}{
    \begin{tabular}{c|lcccccc}
        \toprule
        Dataset & Model & OA & $P_{\mathrm{scd}}$ & $R_{\mathrm{scd}}$ 
        & $F_{\mathrm{scd}}$ & mIoU & SeK \\
        \midrule
        \multirow{6}{*}{HiUCD-Mini}
        & HRSCD4     & 94.03 & 42.31 & 47.68 & 44.83 & 68.04 & 8.67 \\
        & EGMSNet    & 95.02 & 61.78 & 56.35 & 58.94 & 73.79 & 20.37 \\
        & BGSNet     & 92.83 & 69.08 & 43.03 & 53.03 & 69.63 & 16.36 \\
        & CPGNet     & 94.96 & 65.38 & 56.10 & 60.39 & 73.69 & 21.27 \\
        & ChangeMamba & 95.56 & 60.05 & 62.20 & 61.11 & 74.81 & 22.43 \\
        & FoBa       & 94.88 & 47.14 & 57.89 & 51.96 & 70.03 & 13.79 \\
        \midrule
        \multirow{6}{*}{JL1}
        & HRSCD4     & 84.92 & 67.60 & 69.96 & 68.74 & 74.40 & 27.83 \\
        & EGMSNet    & 92.70 & 82.76 & 89.30 & 85.94 & 84.88 & 54.50 \\
        & BGSNet     & 91.82 & 86.71 & 84.34 & 85.51 & 83.11 & 52.21 \\
        & CPGNet     & 93.02 & 83.96 & 89.59 & 86.68 & 85.43 & 55.98 \\
        & ChangeMamba & 94.89 & 89.92 & 91.27 & 90.61 & 89.12 & 65.83 \\
        & FoBa       & 94.22 & 87.77 & 91.05 & 89.40 & 87.54 & 62.02 \\
        \midrule
        \multirow{6}{*}{LandsatSCD}
        & HRSCD4     & 93.19 & 81.78 & 79.06 & 80.37 & 81.44 & 40.95 \\
        & EGMSNet    & 96.98 & 93.99 & 89.14 & 91.50 & 91.06 & 67.00 \\
        & BGSNet     & 93.58 & 93.99 & 74.42 & 83.06 & 83.81 & 47.51 \\
        & CPGNet     & 96.66 & 90.51 & 90.07 & 90.29 & 90.11 & 63.74 \\
        & ChangeMamba & 96.05 & 88.87 & 87.81 & 88.32 & 88.71 & 59.07 \\
        & FoBa       & 95.80 & 87.34 & 88.30 & 87.81 & 87.61 & 56.67 \\
        \midrule
        \multirow{6}{*}{SECOND}
        & HRSCD4     & 84.52 & 51.47 & 52.73 & 52.08 & 67.51 & 12.14 \\
        & EGMSNet    & 87.19 & 59.38 & 62.12 & 60.72 & 71.30 & 19.97 \\
        & BGSNet     & 86.13 & 62.99 & 58.51 & 60.67 & 69.77 & 19.20 \\
        & CPGNet     & 88.32 & 59.69 & 67.22 & 63.22 & 72.55 & 22.35 \\
        & ChangeMamba & 88.74 & 59.20 & 69.59 & 63.97 & 72.97 & 23.30 \\
        & FoBa       & 88.77 & 61.07 & 69.09 & 64.82 & 73.17 & 23.98 \\
        \bottomrule
    \end{tabular}
    }
\end{table}

\begin{table}[!htbp]
\centering
\caption{Robustness under the default white-box setting: joint bitemporal PGD with the semantic-change target. Attack success rates are relative F-score reductions.}
\label{supp:tab:standard_pgd_robustness}
\footnotesize
\setlength{\tabcolsep}{3pt}
\begin{tabular*}{\linewidth}{c|l@{\hspace{6pt}}|@{\extracolsep{\fill}}ccc|ccc|ccc}
\toprule
Dataset & Model
& $F_{\mathrm{bcd}}^{clean}$
& $F_{\mathrm{bcd}}^{adv}$
& $\mathrm{ASR}_{\mathrm{bcd}}$
& $F_{\mathrm{seg}}^{clean}$
& $F_{\mathrm{seg}}^{adv}$
& $\mathrm{ASR}_{\mathrm{seg}}$
& $F_{\mathrm{scd}}^{clean}$
& $F_{\mathrm{scd}}^{adv}$
& $\mathrm{ASR}_{\mathrm{scd}}$ \\
\midrule
\multirow{6}{*}{HiUCD-Mini} & HRSCD4 & 58.06 & 49.65 & 14.48 & 37.53 & 27.02 & 28.01 & 44.83 & 0.91 & 97.97 \\
 & EGMSNet & 69.07 & 57.00 & 17.48 & 61.42 & 41.47 & 32.48 & 58.94 & 1.95 & 96.68 \\
 & BGSNet & 64.25 & 43.84 & 31.77 & 56.48 & 37.77 & 33.13 & 53.03 & 3.83 & 92.78 \\
 & CPGNet & 68.30 & 59.56 & 12.80 & 57.90 & 41.25 & 28.76 & 60.39 & 6.27 & 89.62 \\
 & ChangeMamba & 69.40 & 59.89 & 13.70 & 53.71 & 36.21 & 32.58 & 61.11 & 3.33 & 94.56 \\
 & FoBa & 61.24 & 60.33 & 1.48 & 47.43 & 35.28 & 25.63 & 51.96 & 4.28 & 91.77 \\
\midrule
\multirow{6}{*}{JL1} & HRSCD4 & 78.55 & 52.38 & 33.31 & 41.28 & 1.79 & 95.66 & 68.74 & 0.63 & 99.08 \\
 & EGMSNet & 88.08 & 64.03 & 27.30 & 61.85 & 3.73 & 93.97 & 85.94 & 2.28 & 97.34 \\
 & BGSNet & 86.82 & 61.58 & 29.06 & 59.17 & 4.59 & 92.23 & 85.51 & 9.24 & 89.19 \\
 & CPGNet & 88.55 & 62.16 & 29.81 & 55.66 & 9.11 & 83.64 & 86.68 & 11.89 & 86.29 \\
 & ChangeMamba & 91.75 & 56.81 & 38.08 & 52.45 & 7.80 & 85.14 & 90.61 & 10.08 & 88.88 \\
 & FoBa & 90.42 & 54.19 & 40.07 & 62.22 & 6.73 & 89.19 & 89.40 & 18.06 & 79.79 \\
\midrule
\multirow{6}{*}{LandsatSCD} & HRSCD4 & 82.62 & 66.19 & 19.88 & 26.80 & 16.97 & 36.68 & 80.37 & 35.74 & 55.53 \\
 & EGMSNet & 92.24 & 61.37 & 33.46 & 28.22 & 10.15 & 64.03 & 91.50 & 22.71 & 75.18 \\
 & BGSNet & 85.57 & 55.33 & 35.34 & 39.02 & 7.33 & 81.23 & 83.06 & 13.77 & 83.42 \\
 & CPGNet & 91.31 & 60.90 & 33.30 & 31.49 & 7.67 & 75.63 & 90.29 & 11.14 & 87.66 \\
 & ChangeMamba & 89.97 & 61.47 & 31.68 & 51.30 & 19.44 & 62.10 & 88.32 & 14.40 & 83.69 \\
 & FoBa & 88.87 & 60.67 & 31.73 & 39.50 & 6.21 & 84.29 & 87.81 & 10.71 & 87.81 \\
\midrule
\multirow{6}{*}{SECOND} & HRSCD4 & 66.33 & 43.68 & 34.15 & 10.39 & 3.33 & 67.89 & 52.08 & 5.80 & 88.87 \\
 & EGMSNet & 71.08 & 60.99 & 14.20 & 34.44 & 2.83 & 91.78 & 60.72 & 6.50 & 89.29 \\
 & BGSNet & 69.75 & 48.23 & 30.86 & 23.97 & 3.00 & 87.48 & 60.67 & 5.21 & 91.42 \\
 & CPGNet & 72.31 & 47.30 & 34.59 & 23.88 & 3.60 & 84.95 & 63.22 & 3.30 & 94.77 \\
 & ChangeMamba & 72.67 & 52.91 & 27.20 & 23.53 & 2.81 & 88.05 & 63.97 & 5.36 & 91.62 \\
 & FoBa & 73.06 & 58.06 & 20.53 & 35.25 & 4.77 & 86.46 & 64.82 & 12.04 & 81.43 \\
\bottomrule
\end{tabular*}
\end{table}

\begin{table}[!htbp]
\centering
\caption{Attack success rates for component-level targets under joint bitemporal PGD.}
\label{supp:tab:objective_comparison_diagnostic}
\scriptsize
\setlength{\tabcolsep}{2.2pt}
\renewcommand{\arraystretch}{0.95}
\resizebox{\linewidth}{!}{
\begin{tabular}{c|l|ccc|ccc|ccc|ccc|ccc}
\toprule
\multirow{2}{*}{Dataset}
& \multirow{2}{*}{Model}
& \multicolumn{3}{c|}{Sem.}
& \multicolumn{3}{c|}{BCD}
& \multicolumn{3}{c|}{Sem.+BCD}
& \multicolumn{3}{c|}{SCD}
& \multicolumn{3}{c}{Joint} \\
\cmidrule(lr){3-5}
\cmidrule(lr){6-8}
\cmidrule(lr){9-11}
\cmidrule(lr){12-14}
\cmidrule(lr){15-17}
&
& $\mathrm{ASR}_{\mathrm{bcd}}$
& $\mathrm{ASR}_{\mathrm{seg}}$
& $\mathrm{ASR}_{\mathrm{scd}}$
& $\mathrm{ASR}_{\mathrm{bcd}}$
& $\mathrm{ASR}_{\mathrm{seg}}$
& $\mathrm{ASR}_{\mathrm{scd}}$
& $\mathrm{ASR}_{\mathrm{bcd}}$
& $\mathrm{ASR}_{\mathrm{seg}}$
& $\mathrm{ASR}_{\mathrm{scd}}$
& $\mathrm{ASR}_{\mathrm{bcd}}$
& $\mathrm{ASR}_{\mathrm{seg}}$
& $\mathrm{ASR}_{\mathrm{scd}}$
& $\mathrm{ASR}_{\mathrm{bcd}}$
& $\mathrm{ASR}_{\mathrm{seg}}$
& $\mathrm{ASR}_{\mathrm{scd}}$ \\
\midrule
\multirow{6}{*}{HiUCD-Mini}
& HRSCD4      & 16.50 & 86.69 & 87.50 & 92.48 & 25.12 & 93.85 & 87.13 & 83.85 & 95.69 & 14.48 & 28.01 & 97.97 & 73.29 & 84.35 & 98.19 \\
& EGMSNet     & 61.52 & 75.38 & 90.15 & 98.30 & 53.72 & 98.97 & 66.07 & 75.84 & 91.81 & 17.48 & 32.48 & 96.68 & 58.30 & 77.47 & 95.84 \\
& BGSNet      & 64.64 & 84.04 & 92.53 & 88.22 & 58.76 & 91.01 & 84.92 & 74.29 & 94.22 & 31.77 & 33.13 & 92.78 & 76.77 & 75.44 & 96.01 \\
& CPGNet      & 57.49 & 76.72 & 90.28 & 93.84 & 58.05 & 95.12 & 85.21 & 69.01 & 94.23 & 12.80 & 28.76 & 89.62 & 72.03 & 70.35 & 95.52 \\
& ChangeMamba & 63.94 & 87.29 & 94.94 & 95.72 & 66.22 & 97.64 & 93.28 & 87.57 & 98.05 & 13.70 & 32.58 & 94.56 & 78.87 & 86.98 & 97.52 \\
& FoBa        & 51.13 & 85.55 & 91.89 & 98.34 & 66.28 & 98.96 & 82.65 & 84.07 & 95.00 & 1.48 & 25.63 & 91.77 & 54.15 & 84.35 & 93.99 \\
\midrule
\multirow{6}{*}{JL1}
& HRSCD4      & 32.83 & 78.53 & 65.66 & 83.72 & 52.64 & 88.04 & 68.24 & 78.27 & 77.47 & 33.31 & 95.66 & 99.08 & 58.33 & 83.18 & 86.52 \\
& EGMSNet     & 47.29 & 61.70 & 70.03 & 90.42 & 77.06 & 93.30 & 48.31 & 62.08 & 70.64 & 27.30 & 93.97 & 97.34 & 45.32 & 72.28 & 80.30 \\
& BGSNet      & 85.96 & 91.49 & 96.27 & 77.89 & 69.29 & 78.94 & 93.03 & 90.22 & 95.74 & 29.06 & 92.23 & 89.19 & 71.69 & 90.65 & 90.71 \\
& CPGNet      & 26.28 & 52.53 & 55.36 & 76.74 & 57.62 & 77.78 & 60.87 & 51.50 & 66.46 & 29.81 & 83.64 & 86.29 & 54.25 & 63.85 & 74.93 \\
& ChangeMamba & 59.93 & 69.02 & 76.54 & 79.54 & 61.81 & 83.84 & 82.40 & 66.30 & 85.50 & 38.08 & 85.14 & 88.88 & 74.18 & 72.64 & 85.03 \\
& FoBa        & 39.18 & 67.76 & 67.50 & 88.66 & 79.90 & 90.42 & 69.43 & 72.95 & 78.40 & 40.07 & 89.19 & 79.79 & 59.78 & 81.86 & 79.89 \\

\midrule
\multirow{6}{*}{LandsatSCD}
& HRSCD4      & 3.89 & 22.28 & 23.21 & 32.15 & 3.38 & 34.33 & 23.05 & 18.49 & 34.38 & 19.88 & 36.68 & 55.53 & 25.52 & 32.01 & 51.75 \\
& EGMSNet     & 32.60 & 16.77 & 37.04 & 66.64 & 31.31 & 71.06 & 33.17 & 17.22 & 37.70 & 33.46 & 64.03 & 75.18 & 38.78 & 30.61 & 52.16 \\
& BGSNet      & 11.22 & 26.46 & 20.08 & 49.12 & 24.43 & 53.99 & 18.82 & 26.14 & 26.54 & 35.34 & 81.23 & 83.42 & 22.27 & 47.04 & 42.42 \\
& CPGNet      & 22.23 & 33.90 & 27.62 & 56.23 & 32.46 & 58.91 & 33.39 & 37.30 & 37.29 & 33.30 & 75.63 & 87.66 & 41.22 & 55.80 & 61.86 \\
& ChangeMamba & 19.48 & 58.03 & 30.95 & 60.30 & 32.08 & 64.57 & 31.01 & 58.41 & 40.09 & 31.68 & 62.10 & 83.69 & 31.06 & 66.87 & 60.27 \\
& FoBa        & 52.49 & 53.07 & 66.88 & 65.17 & 40.90 & 69.08 & 58.23 & 54.33 & 70.33 & 31.73 & 84.29 & 87.81 & 53.23 & 71.08 & 82.97 \\
\midrule
\multirow{6}{*}{SECOND}
& HRSCD4      & 20.39 & 46.96 & 59.86 & 72.43 & 15.06 & 77.11 & 66.06 & 42.77 & 77.67 & 34.15 & 67.89 & 88.87 & 57.54 & 57.70 & 88.60 \\
& EGMSNet     & 34.88 & 77.30 & 67.09 & 86.91 & 78.51 & 91.05 & 35.67 & 77.47 & 67.87 & 14.20 & 91.78 & 89.29 & 34.57 & 88.13 & 82.83 \\
& BGSNet      & 37.94 & 68.99 & 66.33 & 73.89 & 53.53 & 81.01 & 58.16 & 63.48 & 73.34 & 30.86 & 87.48 & 91.42 & 41.56 & 77.18 & 82.66 \\
& CPGNet      & 36.71 & 66.40 & 69.55 & 75.32 & 47.03 & 83.85 & 62.51 & 63.41 & 80.00 & 34.59 & 84.95 & 94.77 & 52.62 & 77.91 & 91.34 \\
& ChangeMamba & 42.68 & 80.93 & 85.42 & 83.20 & 61.36 & 87.76 & 81.66 & 80.86 & 91.90 & 27.20 & 88.05 & 91.62 & 36.80 & 83.02 & 88.00 \\
& FoBa        & 44.16 & 78.99 & 72.97 & 88.53 & 84.31 & 91.41 & 59.56 & 80.55 & 79.22 & 20.53 & 86.46 & 81.43 & 44.58 & 83.65 & 82.09 \\

\bottomrule
\end{tabular}
}
\vspace{1mm}
\begin{tablenotes}
\footnotesize
\item Sem., BCD, SCD, and Joint denote attacks optimized using $(\lambda_{sem},\lambda_{bcd},\lambda_{scd})=(1,0,0)$, $(0,1,0)$, $(0,0,1)$, and $(1,1,1)$, respectively. Sem.+BCD corresponds to $(1,1,0)$.
\end{tablenotes}
\end{table}

\begin{table}[!htbp]
    \centering
    \caption{Attack success rates for temporal semantic-change targets under joint bitemporal PGD. The loss targets the $T_1$ output, the $T_2$ output, or both.}
    \label{supp:tab:temporal_target_lscd}
    \footnotesize
    \setlength{\tabcolsep}{1.5pt}
    \renewcommand{\arraystretch}{0.95}
    \begin{tabular*}{\linewidth}{@{\extracolsep{\fill}}c|lc|ccc|ccc|ccc@{}}
        \toprule
        \multirow{2}{*}{Dataset}
        & \multirow{2}{*}{Model}
        & \multirow{2}{*}{Fusion}
        & \multicolumn{3}{c|}{Target $\mathcal{L}_{\mathrm{scd}}^{T_{1}}$}
        & \multicolumn{3}{c|}{Target $\mathcal{L}_{\mathrm{scd}}^{T_{2}}$}
        & \multicolumn{3}{c}{Target $\mathcal{L}_{\mathrm{scd}}$} \\
        \cmidrule(lr){4-6}
        \cmidrule(lr){7-9}
        \cmidrule(lr){10-12}
        &
        &
        & $\mathrm{ASR}_{\mathrm{scd}}^{T_1}$
        & $\mathrm{ASR}_{\mathrm{scd}}^{T_2}$
        & $\mathrm{ASR}_{\mathrm{scd}}$
        & $\mathrm{ASR}_{\mathrm{scd}}^{T_1}$
        & $\mathrm{ASR}_{\mathrm{scd}}^{T_2}$
        & $\mathrm{ASR}_{\mathrm{scd}}$
        & $\mathrm{ASR}_{\mathrm{scd}}^{T_1}$
        & $\mathrm{ASR}_{\mathrm{scd}}^{T_2}$
        & $\mathrm{ASR}_{\mathrm{scd}}$ \\
        \midrule
        \multirow{6}{*}{HiUCD-Mini}
        & HRSCD4      & Concat.           & 99.36 & 34.19 & 69.91 & 38.20 & 98.62 & 65.54 & 98.14 & 97.76 & 97.97 \\
        & EGMSNet     & Diff.             & 98.04 & 61.20 & 81.38 & 62.46 & 97.37 & 78.26 & 96.77 & 96.60 & 96.68 \\
        & BGSNet      & Cross-Attn.+Diff. & 96.55 & 60.03 & 80.11 & 60.86 & 96.43 & 76.89 & 92.11 & 93.65 & 92.78 \\
        & CPGNet      & Diff.+Concat.     & 96.43 & 64.06 & 81.27 & 53.69 & 95.07 & 73.09 & 88.88 & 90.51 & 89.62 \\
        & ChangeMamba & SSM+Concat.       & 98.35 & 73.25 & 86.56 & 75.56 & 98.13 & 86.17 & 94.84 & 94.17 & 94.56 \\
        & FoBa        & Mul.+Concat.      & 97.07 & 59.34 & 78.77 & 63.64 & 94.93 & 78.81 & 92.88 & 90.62 & 91.77 \\

        \midrule
        \multirow{6}{*}{JL1}
        & HRSCD4      & Concat.           & 99.68 & 58.25 & 78.92 & 55.04 & 99.58 & 77.36 & 99.21 & 99.03 & 99.08 \\
        & EGMSNet     & Diff.             & 99.43 & 49.45 & 74.50 & 52.84 & 98.92 & 75.83 & 98.59 & 96.38 & 97.34 \\
        & BGSNet      & Cross-Attn.+Diff. & 99.44 & 79.23 & 89.37 & 75.23 & 97.83 & 86.49 & 89.69 & 88.92 & 89.19 \\
        & CPGNet      & Diff.+Concat.     & 98.30 & 78.63 & 88.48 & 68.70 & 96.76 & 82.72 & 85.01 & 88.23 & 86.29 \\
        & ChangeMamba & SSM+Concat.       & 99.36 & 68.76 & 84.08 & 71.22 & 98.05 & 84.62 & 91.00 & 86.98 & 88.88 \\
        & FoBa        & Mul.+Concat.      & 98.94 & 73.17 & 86.08 & 71.79 & 97.64 & 84.69 & 79.62 & 79.52 & 79.79 \\

        \midrule
        \multirow{6}{*}{LandsatSCD}
        & HRSCD4      & Concat.           & 64.07 & 23.95 & 44.03 & 20.42 & 65.43 & 42.90 & 54.21 & 56.68 & 55.53 \\
        & EGMSNet     & Diff.             & 72.79 & 59.96 & 66.38 & 55.29 & 68.46 & 61.87 & 75.90 & 74.64 & 75.18 \\
        & BGSNet      & Cross-Attn.+Diff. & 80.38 & 72.80 & 76.60 & 68.49 & 87.39 & 77.92 & 81.08 & 85.84 & 83.42 \\
        & CPGNet      & Diff.+Concat.     & 89.12 & 77.70 & 83.41 & 73.13 & 89.93 & 81.53 & 85.64 & 89.57 & 87.66 \\
        & ChangeMamba & SSM+Concat.       & 88.07 & 66.03 & 77.08 & 56.49 & 88.91 & 72.66 & 81.49 & 85.76 & 83.69 \\
        & FoBa        & Mul.+Concat.      & 90.51 & 85.11 & 87.81 & 78.61 & 90.74 & 84.67 & 86.91 & 88.72 & 87.81 \\

        \midrule
        \multirow{6}{*}{SECOND}
        & HRSCD4      & Concat.           & 97.51 & 50.88 & 73.18 & 60.91 & 95.71 & 79.06 & 93.56 & 84.55 & 88.87 \\
        & EGMSNet     & Diff.             & 95.78 & 50.19 & 71.96 & 59.03 & 92.74 & 76.64 & 92.59 & 86.33 & 89.29 \\
        & BGSNet      & Cross-Attn.+Diff. & 95.39 & 68.00 & 81.20 & 79.08 & 94.13 & 86.88 & 93.61 & 89.48 & 91.42 \\
        & CPGNet      & Diff.+Concat.     & 97.22 & 65.16 & 80.59 & 80.67 & 97.23 & 89.26 & 96.04 & 93.59 & 94.77 \\
        & ChangeMamba & SSM+Concat.       & 96.49 & 73.50 & 84.54 & 75.29 & 94.98 & 85.53 & 92.85 & 90.42 & 91.62 \\
        & FoBa        & Mul.+Concat.      & 92.80 & 55.53 & 73.47 & 59.72 & 90.73 & 75.80 & 86.92 & 76.39 & 81.43 \\
        \bottomrule
    \end{tabular*}
\end{table}

\begin{table}[!htbp]
    \centering
    \caption{Attack success rates for PGD access to $T_1$, $T_2$, or both inputs, using the semantic-change target.}
    \label{supp:tab:cross_temporal_model_extended}
    \scriptsize
    \setlength{\tabcolsep}{1.5pt}
    \renewcommand{\arraystretch}{0.95}
    \begin{tabular*}{\linewidth}{@{\extracolsep{\fill}}c|lc|ccc|ccc|ccc@{}}
        \toprule
        \multirow{2}{*}{Dataset}
        & \multirow{2}{*}{Model}
        & \multirow{2}{*}{Fusion}
        & \multicolumn{3}{c|}{$T_1$ Perturbed}
        & \multicolumn{3}{c|}{$T_2$ Perturbed}
        & \multicolumn{3}{c}{$T_1+T_2$ Perturbed} \\
        \cmidrule(lr){4-6}
        \cmidrule(lr){7-9}
        \cmidrule(lr){10-12}
        &
        &
        & $\mathrm{ASR}_{\mathrm{scd}}^{T_1}$
        & $\mathrm{ASR}_{\mathrm{scd}}^{T_2}$
        & $\mathrm{ASR}_{\mathrm{scd}}$
        & $\mathrm{ASR}_{\mathrm{scd}}^{T_1}$
        & $\mathrm{ASR}_{\mathrm{scd}}^{T_2}$
        & $\mathrm{ASR}_{\mathrm{scd}}$
        & $\mathrm{ASR}_{\mathrm{scd}}^{T_1}$
        & $\mathrm{ASR}_{\mathrm{scd}}^{T_2}$
        & $\mathrm{ASR}_{\mathrm{scd}}$ \\
        \midrule
        \multirow{6}{*}{HiUCD-Mini}
        & HRSCD4      & Concat.            & 99.02 & 46.43 & 75.36 & 36.82 & 97.85 & 64.30 & 98.10 & 97.81 & 97.97 \\
        & EGMSNet     & Diff.              & 96.19 & 21.68 & 62.54 & 21.71 & 95.16 & 54.90 & 96.77 & 96.58 & 96.68 \\
        & BGSNet      & Cross-Attn.+Diff.  & 90.84 & 42.68 & 69.18 & 33.76 & 92.08 & 59.99 & 92.19 & 93.50 & 92.78 \\
        & CPGNet      & Diff.+Concat.      & 91.49 & 32.19 & 63.75 & 21.11 & 90.38 & 53.52 & 88.94 & 90.38 & 89.62 \\
        & ChangeMamba & SSM+Concat.        & 95.18 & 30.43 & 64.85 & 18.88 & 95.60 & 54.81 & 94.81 & 94.27 & 94.56 \\
        & FoBa        & Mul.+Concat.       & 90.15 & 58.87 & 75.01 & 63.87 & 87.11 & 75.12 & 92.81 & 90.66 & 91.77 \\
    
        \midrule
        
        \multirow{6}{*}{JL1}
        & HRSCD4      & Concat.            & 99.60 & 45.48 & 72.47 & 45.68 & 99.35 & 72.58 & 99.14 & 99.03 & 99.08 \\
        & EGMSNet     & Diff.              & 99.18 & 27.98 & 63.66 & 29.36 & 97.89 & 63.55 & 98.49 & 96.19 & 97.34 \\
        & BGSNet      & Cross-Attn.+Diff.  & 82.83 & 76.34 & 79.60 & 70.67 & 89.50 & 80.05 & 89.71 & 88.67 & 89.19 \\
        & CPGNet      & Diff.+Concat.      & 93.36 & 46.23 & 69.82 & 31.66 & 92.31 & 61.95 & 85.18 & 87.39 & 86.29 \\
        & ChangeMamba & SSM+Concat.        & 98.12 & 34.80 & 66.50 & 41.18 & 96.38 & 68.75 & 91.12 & 86.63 & 88.88 \\
        & FoBa        & Mul.+Concat.       & 88.03 & 71.39 & 79.73 & 70.60 & 89.93 & 80.24 & 79.91 & 79.68 & 79.79 \\
        \midrule
        
        \multirow{6}{*}{LandsatSCD}
        & HRSCD4      & Concat.            & 48.20 & 12.45 & 30.34 & 12.25 & 52.06 & 32.13 & 54.22 & 56.83 & 55.53 \\
        & EGMSNet     & Diff.              & 63.66 & 42.25 & 52.96 & 22.68 & 47.06 & 34.86 & 75.79 & 74.57 & 75.18 \\
        & BGSNet      & Cross-Attn.+Diff.  & 69.62 & 50.01 & 59.83 & 45.18 & 81.80 & 63.45 & 81.03 & 85.82 & 83.42 \\
        & CPGNet      & Diff.+Concat.      & 81.62 & 62.28 & 71.96 & 46.32 & 78.94 & 62.62 & 85.70 & 89.63 & 87.66 \\
        & ChangeMamba & SSM+Concat.        & 73.85 & 34.13 & 54.04 & 23.20 & 75.94 & 49.50 & 81.60 & 85.80 & 83.69 \\
        & FoBa        & Mul.+Concat.       & 83.37 & 65.65 & 74.52 & 42.80 & 74.46 & 58.62 & 86.90 & 88.72 & 87.81 \\

        \midrule
        
        \multirow{6}{*}{SECOND}
        & HRSCD4      & Concat.            & 93.78 & 41.14 & 66.32 & 43.36 & 82.51 & 63.79 & 93.54 & 84.59 & 88.87 \\
        & EGMSNet     & Diff.              & 94.02 & 27.90 & 59.47 & 25.80 & 89.08 & 58.86 & 92.61 & 86.26 & 89.29 \\
        & BGSNet      & Cross-Attn.+Diff.  & 92.51 & 39.78 & 65.19 & 46.34 & 89.30 & 68.60 & 93.57 & 89.42 & 91.42 \\
        & CPGNet      & Diff.+Concat.      & 95.97 & 42.62 & 68.30 & 64.69 & 94.85 & 80.33 & 96.06 & 93.58 & 94.77 \\
        & ChangeMamba & SSM+Concat.        & 93.59 & 28.07 & 59.53 & 26.13 & 89.65 & 59.15 & 92.89 & 90.45 & 91.62 \\
        & FoBa        & Mul.+Concat.       & 85.78 & 40.79 & 62.45 & 53.96 & 77.48 & 66.16 & 86.81 & 76.43 & 81.43 \\
        \bottomrule
    \end{tabular*}
\end{table}

\begin{table}[!htbp]
\centering
\caption{$\mathrm{ASR}_{\mathrm{scd}}$ (\%) for perturbation methods. All methods are applied to both temporal inputs; gradient-based attacks use the semantic-change target.}
\label{supp:tab:attack_method_asr_with_pgd}
\resizebox{.7\linewidth}{!}{
\begin{tabular}{c|l|cccccc}
\toprule
Dataset & Model
& Gaussian
& Jitter
& FGSM
& I-FGSM
& C\&W
& PGD \\
\midrule
\multirow{6}{*}{HiUCD-Mini}
& HRSCD4 & 0.07 & 0.23 & 75.48 & 98.00 & 97.58 & 97.97 \\
& EGMSNet & 0.02 & 2.18 & 78.66 & 96.61 & 95.82 & 96.68 \\
& BGSNet & 0.01 & 0.00 & 63.27 & 93.61 & 91.42 & 92.78 \\
& CPGNet & 0.06 & 0.84 & 54.98 & 90.98 & 88.82 & 89.62 \\
& ChangeMamba & 0.04 & 1.23 & 62.37 & 95.54 & 90.79 & 94.56 \\
& FoBa & 0.00 & 0.59 & 73.35 & 92.27 & 90.17 & 91.77 \\
\midrule
\multirow{6}{*}{JL1}
& HRSCD4 & 0.00 & 1.37 & 58.40 & 99.19 & 90.20 & 99.08 \\
& EGMSNet & 0.00 & 2.18 & 63.69 & 97.85 & 79.83 & 97.34 \\
& BGSNet & 0.00 & 1.20 & 46.46 & 89.00 & 66.50 & 89.19 \\
& CPGNet & 0.01 & 1.27 & 35.73 & 88.25 & 60.35 & 86.29 \\
& ChangeMamba & 0.10 & 0.86 & 35.28 & 90.38 & 55.52 & 88.88 \\
& FoBa & 0.04 & 0.85 & 37.03 & 80.12 & 62.33 & 79.79 \\
\midrule
\multirow{6}{*}{LandsatSCD}
& HRSCD4 & 0.01 & 4.00 & 18.70 & 59.01 & 37.37 & 55.53 \\
& EGMSNet & 0.03 & 7.96 & 28.51 & 78.07 & 43.42 & 75.18 \\
& BGSNet & 0.06 & 6.13 & 26.38 & 84.37 & 70.72 & 83.42 \\
& CPGNet & 0.11 & 7.19 & 35.95 & 89.44 & 68.52 & 87.66 \\
& ChangeMamba & 0.04 & 5.71 & 31.43 & 85.94 & 53.20 & 83.69 \\
& FoBa & 0.03 & 4.31 & 34.82 & 89.04 & 64.67 & 87.81 \\
\midrule
\multirow{6}{*}{SECOND}
& HRSCD4 & 0.01 & 0.04 & 41.87 & 89.62 & 72.74 & 88.87 \\
& EGMSNet & 0.01 & 0.30 & 68.27 & 90.10 & 71.64 & 89.29 \\
& BGSNet & 0.00 & 0.45 & 63.07 & 92.02 & 75.74 & 91.42 \\
& CPGNet & 0.00 & 0.22 & 67.52 & 95.14 & 74.35 & 94.77 \\
& ChangeMamba & 0.00 & 0.46 & 60.64 & 91.94 & 70.32 & 91.62 \\
& FoBa & 0.00 & 0.43 & 53.00 & 82.23 & 56.91 & 81.43 \\
\bottomrule
\end{tabular}
}
\end{table}

\FloatBarrier
\clearpage

\begin{table}[!htbp]
    \centering
    \caption{Cross-architecture PGD transferability with the semantic-change target. Rows are source models, columns are target models, and entries are $\mathrm{ASR}_{\mathrm{scd}}$ values. Diagonal entries are white-box attacks noted in \textbf{bold}.}
    \label{supp:tab:cross_architecture_transfer_datasets}
    \footnotesize
    \setlength{\tabcolsep}{2.5pt}
    \renewcommand{\arraystretch}{0.95}
    \begin{tabular*}{\linewidth}{@{\extracolsep{\fill}}c|ll|cccccc@{}}
        \toprule
        \multirow{2}{*}{Dataset}
        & \multirow{2}{*}{Source}
        & \multirow{2}{*}{Family}
        & \multicolumn{6}{c}{Target model} \\
        &
        &
        & HRSCD4
        & EGMSNet
        & BGSNet
        & CPGNet
        & ChangeMamba
        & FoBa \\
        \midrule

        \multirow{6}{*}{HiUCD-Mini}
        & HRSCD4      & CNN         & \textbf{97.97} & 0.20 & 0.39 & 0.28 & 0.22 & 0.16 \\
        & EGMSNet     & CNN         & 1.14 & \textbf{96.68} & 3.90 & 4.21 & 4.61 & 4.40 \\
        & BGSNet      & Transformer & 0.74 & 3.10 & \textbf{92.78} & 19.51 & 5.32 & 5.21 \\
        & CPGNet      & Transformer & 1.17 & 3.83 & 27.74 & \textbf{89.62} & 8.28 & 8.03 \\
        & ChangeMamba & Mamba       & 1.08 & 4.36 & 7.16 & 9.12 & \textbf{94.56} & 18.98 \\
        & FoBa        & Mamba       & 0.78 & 3.12 & 6.46 & 5.65 & 16.66 & \textbf{91.77} \\
        \midrule

        \multirow{6}{*}{JL1}
        & HRSCD4      & CNN         & \textbf{99.08} & 0.34 & 0.14 & 0.48 & 0.81 & 0.32 \\
        & EGMSNet     & CNN         & 0.61 & \textbf{97.34} & 1.13 & 1.50 & 1.92 & 1.23 \\
        & BGSNet      & Transformer & 0.40 & 1.55 & \textbf{89.19} & 12.77 & 3.13 & 2.16 \\
        & CPGNet      & Transformer & 0.36 & 1.67 & 17.14 & \textbf{86.29} & 4.02 & 2.80 \\
        & ChangeMamba & Mamba       & 0.38 & 1.63 & 3.08 & 3.51 & \textbf{88.88} & 5.54 \\
        & FoBa        & Mamba       & 0.52 & 1.72 & 3.08 & 3.38 & 9.06 & \textbf{79.79} \\
        \midrule

        \multirow{6}{*}{LandsatSCD}
        & HRSCD4      & CNN         & \textbf{55.53} & 0.89 & 0.87 & 1.58 & 0.90 & 0.92 \\
        & EGMSNet     & CNN         & 1.60 & \textbf{75.18} & 2.09 & 4.39 & 3.21 & 3.36 \\
        & BGSNet      & Transformer & 0.66 & 1.08 & \textbf{83.42} & 5.47 & 1.77 & 1.64 \\
        & CPGNet      & Transformer & 0.97 & 1.69 & 8.21 & \textbf{87.66} & 3.24 & 2.92 \\
        & ChangeMamba & Mamba       & 0.96 & 2.17 & 2.93 & 5.09 & \textbf{83.69} & 6.23 \\
        & FoBa        & Mamba       & 1.44 & 2.95 & 3.69 & 6.23 & 8.52 & \textbf{87.81} \\
        \midrule

        \multirow{6}{*}{SECOND}
        & HRSCD4      & CNN         & \textbf{88.87} & 1.05 & 0.72 & 0.86 & 0.97 & 0.95 \\
        & EGMSNet     & CNN         & 0.85 & \textbf{89.29} & 3.75 & 3.78 & 4.83 & 4.41 \\
        & BGSNet      & Transformer & 0.57 & 2.01 & \textbf{91.42} & 13.28 & 3.96 & 3.18 \\
        & CPGNet      & Transformer & 0.97 & 6.50 & 53.28 & \textbf{94.77} & 13.60 & 10.96 \\
        & ChangeMamba & Mamba       & 0.63 & 3.52 & 5.69 & 5.54 & \textbf{91.62} & 10.61 \\
        & FoBa        & Mamba       & 0.62 & 3.23 & 4.77 & 5.00 & 11.90 & \textbf{81.43} \\
        \bottomrule
    \end{tabular*}
\end{table}

\FloatBarrier
\clearpage

\begin{table}[!htbp]
    \centering
    \caption{Class-wise vulnerability (\%) under joint bitemporal PGD with the semantic-change target. The first row gives each class's source-or-target frequency among changed transitions; model entries give $F_{\mathrm{scd}}^{c,\mathrm{clean}} / \mathrm{ASR}_{\mathrm{scd}}^{c}$. HiUCD-Mini frequencies do not sum to 100\% because about 3\% of changed transitions involve the omitted unlabeled class.}
    \label{supp:tab:class_wise_vulnerability}

    \begin{minipage}{\textwidth}
        \centering
        \scriptsize
        \setlength{\tabcolsep}{3.2pt}
        \renewcommand{\arraystretch}{0.92}
        \textit{(a) HiUCD-Mini}\par
        \resizebox{\textwidth}{!}{
        \begin{tabular}{p{2.2cm}|*{9}{>{\centering\arraybackslash}p{1.45cm}}}
            \toprule
            \textbf{Model / Stat.} & \textbf{Water} & \textbf{Grass} & \textbf{Building} & \textbf{Green House} & \textbf{Road} & \textbf{Bridge} & \textbf{Others} & \textbf{Bareland} & \textbf{Woodland} \\
            \midrule
            Frequency   & 0.44 & 20.87 & 11.74 & 0.23 & 23.89 & 0.45 & 12.31 & 26.40 & 1.19 \\
            \midrule
            HRSCD4      & 0.00/0.00 & 49.11/14.94 & 77.02/13.25 & 0.00/0.00 & 70.44/11.95 & 0.00/0.00 & 49.80/95.15 & 62.72/35.35 & 66.21/14.52 \\
            EGMSNet     & 60.62/48.89 & 49.06/14.02 & 83.38/8.25 & 69.38/80.68 & 73.35/12.04 & 74.14/7.42 & 63.79/89.71 & 66.72/32.25 & 73.71/9.50 \\
            BGSNet      & 64.62/52.69 & 52.58/13.13 & 78.21/8.36 & 69.13/77.96 & 73.89/11.00 & 15.89/2.43 & 64.17/76.64 & 70.11/31.84 & 76.23/7.51 \\
            CPGNet      & 67.04/41.20 & 51.13/11.34 & 80.07/6.63 & 74.78/58.10 & 74.79/8.96 & 15.99/5.35 & 65.47/77.67 & 71.04/29.17 & 78.72/6.65 \\
            ChangeMamba & 65.95/41.95 & 56.43/13.25 & 74.33/10.26 & 58.80/73.42 & 73.83/11.45 & 5.81/16.49 & 60.57/89.40 & 67.12/31.95 & 74.31/5.48 \\
            FoBa        & 50.84/27.34 & 57.00/14.37 & 74.96/8.94 & 37.97/50.70 & 71.09/9.93 & 0.05/0.00 & 52.13/90.97 & 61.13/25.98 & 69.18/4.65 \\
            \bottomrule
        \end{tabular}
        }

    \vspace{1.2mm}

        \textit{(b) JL1}\par
        \resizebox{\textwidth}{!}{
        \begin{tabular}{p{2.2cm}|*{5}{>{\centering\arraybackslash}p{2.25cm}}}
            \toprule
            \textbf{Model / Stat.} & \textbf{Cropland} & \textbf{N.v.g. Surface} & \textbf{Road} & \textbf{Building} & \textbf{Forest} \\
            \midrule
            Frequency   & 50.00 & 21.48 & 3.93 & 3.28 & 21.32 \\
            \midrule
            HRSCD4      & 30.32/96.19 & 68.92/98.18 & 53.25/99.62 & 43.57/81.87 & 51.62/99.55 \\
            EGMSNet     & 28.98/94.31 & 86.97/95.52 & 89.87/100.00 & 82.70/81.06 & 82.56/98.60 \\
            BGSNet      & 29.49/46.73 & 87.14/99.59 & 77.66/99.98 & 77.24/86.33 & 83.46/98.89 \\
            CPGNet      & 29.57/42.63 & 87.41/92.93 & 66.45/99.67 & 68.91/61.54 & 81.59/94.15 \\
            ChangeMamba & 30.89/40.72 & 83.46/94.86 & 55.22/97.69 & 62.81/69.82 & 82.34/95.21 \\
            FoBa        & 29.79/17.39 & 87.62/98.63 & 88.49/100.00 & 81.87/84.50 & 85.57/97.81 \\
            \bottomrule
        \end{tabular}
        }

    \vspace{1.2mm}

        \textit{(c) LandsatSCD}\par
        \resizebox{\textwidth}{!}{
        \begin{tabular}{p{2.2cm}|*{4}{>{\centering\arraybackslash}p{2.85cm}}}
            \toprule 
            \textbf{Model / Stat.} & \textbf{Farmland} & \textbf{Desert} & \textbf{Building} & \textbf{Water} \\
            \midrule
            Frequency   & 39.03 & 46.22 & 5.93 & 8.82 \\
            \midrule
            HRSCD4      & 31.76/36.33 & 45.22/33.04 & 5.30/72.81 & 51.73/36.37 \\
            EGMSNet     & 53.23/61.94 & 24.34/62.96 & 10.84/61.05 & 52.71/67.26 \\
            BGSNet      & 56.00/75.34 & 19.11/56.96 & 68.10/82.89 & 51.90/94.33 \\
            CPGNet      & 43.80/71.34 & 33.40/66.29 & 71.01/82.28 & 9.23/78.69 \\
            ChangeMamba & 55.67/79.73 & 28.85/66.45 & 69.78/69.14 & 51.50/76.00 \\
            FoBa        & 57.34/83.61 & 19.48/72.57 & 68.12/84.30 & 52.57/89.37 \\
            \bottomrule
        \end{tabular}
        }

    \vspace{1.2mm}

        \textit{(d) SECOND}\par
        \resizebox{\textwidth}{!}{
        \begin{tabular}{p{2.2cm}|*{6}{>{\centering\arraybackslash}p{1.90cm}}}
            \toprule
            \textbf{Model / Stat.} & \textbf{Water} & \textbf{N.v.g. Surface} & \textbf{Low Vegetation} & \textbf{Tree} & \textbf{Buildings} & \textbf{Sports Field} \\ 
            \midrule
            Frequency   & 1.38 & 37.61 & 25.86 & 8.17 & 26.34 & 0.64 \\
            \midrule
            HRSCD4      & 0.00/0.00 & 24.83/55.07 & 24.55/87.44 & 0.00/0.00 & 23.32/60.96 & 0.00/0.00 \\
            EGMSNet     & 47.11/99.73 & 18.18/82.84 & 48.78/96.34 & 37.79/94.57 & 30.42/64.92 & 58.82/96.49 \\
            BGSNet      & 22.43/96.68 & 24.37/86.20 & 37.47/94.80 & 19.91/92.75 & 25.06/66.83 & 38.55/86.52 \\
            CPGNet      & 22.15/95.76 & 25.64/80.33 & 36.80/89.13 & 20.10/82.75 & 25.39/67.09 & 37.11/90.94 \\
            ChangeMamba & 22.71/97.67 & 25.69/83.79 & 35.00/92.10 & 18.02/90.67 & 26.62/69.06 & 36.69/93.72 \\
            FoBa        & 46.14/96.91 & 23.81/73.25 & 49.11/93.77 & 35.35/90.91 & 24.34/28.95 & 68.03/96.99 \\
            \bottomrule
        \end{tabular}
        }
    \end{minipage}
\end{table}


\end{document}